\documentclass[sigplan,screen]{acmart}
\usepackage{amsfonts}
\usepackage{amsmath}
\usepackage{bbding}
\usepackage{color}
\usepackage{comment}
\usepackage{epsfig} 
\usepackage{graphics}
\usepackage{lipsum}
\usepackage[caption=false]{subfig}
\usepackage{multirow}
\usepackage{times}
\usepackage{tcolorbox}
\usepackage{url}
\usepackage{wrapfig}
\usepackage{svg}
\usepackage[ruled,vlined]{algorithm2e}

\usepackage{tikz}
\usetikzlibrary{patterns}
\usetikzlibrary{positioning, intersections, arrows, shapes.geometric, calc, shapes.callouts, decorations.pathreplacing, decorations.markings, shapes}
\usepackage{pifont}
\usepackage{pgfplots}
\usepgfplotslibrary{statistics}
\usepackage{pgfplotstable}

\newcommand{\eat}[1]{}

\copyrightyear{2023} 
\acmYear{2023} 
\setcopyright{rightsretained} 
\acmConference[CHASE '23]{ACM/IEEE International Conference on Connected Health: Applications, Systems and Engineering Technologies}{June 21--23, 2023}{Orlando, FL, USA}
\acmBooktitle{ACM/IEEE International Conference on Connected Health: Applications, Systems and Engineering Technologies (CHASE '23), June 21--23, 2023, Orlando, FL, USA}
\acmDOI{10.1145/3580252.3586971}
\acmISBN{979-8-4007-0102-3/23/06}

\begin{document}
\title[Efficient and Direct Inference of HRV using Both Signal Processing and ML]{Efficient and Direct Inference of Heart Rate Variability using Both Signal Processing and Machine Learning}


\author{Yuntong Zhang}
\affiliation{%
 \institution{University of Texas at San Antonio}
 \country{USA}
}

\author{Jingye Xu}
\affiliation{%
 \institution{University of Texas at San Antonio}
 \country{USA}
}

\author{Mimi Xie}
\affiliation{%
 \institution{University of Texas at San Antonio}
 \country{USA}
}

\author{Dakai Zhu}
\affiliation{%
 \institution{University of Texas at San Antonio}
 \country{USA}
}

\author{Houbing Song}
\affiliation{%
 \institution{University Of Maryland}
 \country{USA}
}
\author{Wei Wang}
\affiliation{%
 \institution{University of Texas at San Antonio}
 \country{USA}
}
  

\renewcommand{\shortauthors}{Zhang et al.}
\begin{abstract}

Heart Rate Variability (HRV) measures the variation of the time between consecutive heartbeats 
and is a major indicator of physical and mental health. Recent research has demonstrated that photoplethysmography (PPG) sensors can be used to infer HRV. However, many prior studies had high errors because they only employed signal processing or machine learning (ML), or because they indirectly
inferred HRV, 
or because there lacks large training datasets. 
Many prior studies may also require large ML models. 
The low accuracy and large model sizes limit their applications to small embedded devices and potential future use in healthcare.


To address the above issues, we first collected a large dataset of PPG signals and HRV ground truth. 
With this dataset, we developed HRV models that combine signal processing and ML to directly infer HRV. Evaluation results show that our method had errors between $3.5\%$ to $25.7\%$ and outperformed signal-processing-only and ML-only methods. We also explored different ML models, which showed that Decision Trees and Multi-level Perceptrons have $13.0\%$ and $9.1\%$ errors on average with models at most hundreds of KB and inference time less than 1ms. Hence, they are more suitable for small embedded devices and potentially enable the future use of PPG-based HRV monitoring in healthcare.

\end{abstract}

\keywords{Heart Rate Variability, Machine Learning, Photoplethysmography, Signal Processing}
\maketitle



\section{Introduction}
Heart Rate Variability (HRV) measures the variation of the time intervals of consecutive heartbeats and is a major indicator for health conditions, such as coronary artery disease, heart failure, hyperlipidemia, and hypertension~\cite{2012-Xhyheri-HRVToday}. HRV is traditionally measured using electrocardiographic (ECG) devices, 
which record the heart's rhythm. However, ECGs 
can be expensive and require attaching several electrodes to human bodies, which can be inconvenient to use.

As an alternative to ECG, Photoplethysmography (PPG) sensors, which monitor light signal changes in blood flows, can also be utilized to measure heart rate (HR) and HRV~\cite{2007-Wang-TBCS-PPGEar,zhang2014troika}. A PPG sensor can be placed on the human skin to provide HR/HRV readings and is more convenient to use. Many studies demonstrated the potential of PPG sensors in HRV monitoring~\cite{everson2019biotranslator,wittenberg2020evaluation}. However, these studies usually face the following four limitations, restraining their application to small embedded devices and healthcare in the future.

First, some of these studies relied on only signal processing for HRV estimation~\cite{zhang2014troika,bhowmik2017novel}, which may negatively affect the inference accuracy. PPG sensor signals typically suffer from large noises, particularly, the noises from motion artifacts (MA). To infer HRV with good accuracy, these noises must be removed or reduced. Although signal processing techniques can remove many noises (including MA), these relatively static techniques may not be able to handle all types of noises, leading to low HRV accuracy.


Second, many studies also only employed machine learning (ML) techniques to infer HRV~\cite{everson2019biotranslator,chiu2020reconstructing}.
Although many ML models by themselves are sophisticated enough to handle all types of noises, the resulting models can be too large and/or too slow for small embedded devices. Moreover, training and tuning ML models with raw PPG data can be quite challenging. ML models without enough tuning may also have low accuracy.

Third, some prior studies also focused on the inference of the interval lengths between consecutive heart beats~\cite{everson2019biotranslator,xu2019deep}, which are known as RR intervals~\cite{dohare2014efficient,GOLDBERGER201811}. That is, these studies do not direct infer HRV metrics, such as SDNN or RMSSD (more in Section~\ref{sec:background}). This indirect inference strictly follows HRV's definition. However, due to error amplification, this indirection can significantly increase the errors of the final HRV estimations.

Fourth, many prior studies also relied on small datasets. For example, a popular PPG dataset, the IEEE Signal Processing Cup (ISPC) dataset, has PPG signals recording lasted for only five minutes~\cite{zhang2014troika}. However, a typical HRV inference requires an ECG monitoring length of half to five minutes~\cite{acharya2006heart}. Hence, it is difficult to conduct HRV inference studies with short recordings, especially for studies with neural networks.

We have studied ML-based HR estimation using PPG in our prior work \cite{yuntong2022hr}. In this paper, our goal is to design an HRV inference methodology that can provide high accuracy for resource-constrained embedded devices. To achieve this goal, we designed a \textit{compound and direct} method that combines signal processing and ML to directly infer HRV (i.e., RMSSD/SDNN). More specifically, we first employed signal processing to remove outliers and noises from raw PPG signals and convert the PPG signals into rough HR readings and HRV readings. These rough HR/HRV readings are then fed into an ML model to infer RMSSD/SDNN. Applying ML after signal processing provides additional/better noise removal, and hence, more accurate HRV estimations. Applying signal processing before ML avoids the need for large and slow ML models. 
Moreover, the direct inference of RMSSD/SDNN, instead of inferring RR intervals as a proxy, 
further improves accuracy.

To explore the impact of various ML algorithms, 
we also evaluated 
different ML models, 
including Decision Tree (DT), Random Forest (RF), K-nearest neighbor (KNN), Support vector machines (SVM), and Multi-layer perceptron (MLP). To provide more reliable results, we also collected a new dataset of PPG signals, with ECG readings as ground truth. This dataset contains three 2-hour-long PPG/ECG traces for one human subject performing different activities, including office work, sleeping, and sitting.

Evaluation results show that our compound and direct method has 3.5\% to 25.7\% errors for various activities and monitoring lengths. Both the lowest 3.5\% error for RMSSD and the lowest 5.1\% error for SDNN were obtained with a monitoring length of 300 seconds per HRV estimation. Note that, 300-second HRV monitoring can be used for caring for chronic renal failure and diabetes~\cite{acharya2006heart}, showing the healthcare potential of PPG-based HRV monitoring. The evaluation results also show that our method is significantly more accurate than the signal-processing-only and ML-only methods.

Moreover, the model exploration showed that 
DT and MLP models are usually smaller with good accuracy, making them more suitable for small embedded devices. DT models have an average error of 13.0\% and are usually less than 10KB with inference time less than 10$\mu$s. MLP models have an average error of 9.1\% and are less than 469KB with inference time less than 1ms.  
These results corroborate the "Rashomon" theory~\cite{Rashomon-ML} that, for some problems, there exist simple models with good accuracy and meet special requirements, such as the limited memory size of embedded devices in this case.



The contributions of this paper include,
\vspace{-2mm}
\begin{itemize}
    \item The compound and direct HRV inference methodology combines signal processing and ML to directly infer RMSSD/SDNN to achieve high accuracy with small and fast ML models.

    \item A systematic exploration of different ML algorithms to study their impact on the accuracy, model size, and time on HRV inference. This exploration showed that Decision Trees and MLP can achieve high accuracy with small/fast models suitable for embedded devices.

    \item A comprehensive PPG/ECG dataset to study HR/HRV inference, which contains traces of different activity intensities lasting for 2 hours. 
\end{itemize}

The rest of the paper is structured as follows.
Section~\ref{sec:background} discusses the background on HRV and the motivation of our work. 
Section~\ref{sec:method} presents our compound and direct method.
Section~\ref{sec:Evaluation} presents evaluation results. 
Section \ref{sec:related_work} discusses related work, and section~\ref{sec:Conclusion} concludes the paper.

\section{Background and Motivation}\label{sec:background}

\subsection{HRV from ECG}

\begin{figure}
  \centering
  \includegraphics[width=0.8\linewidth]{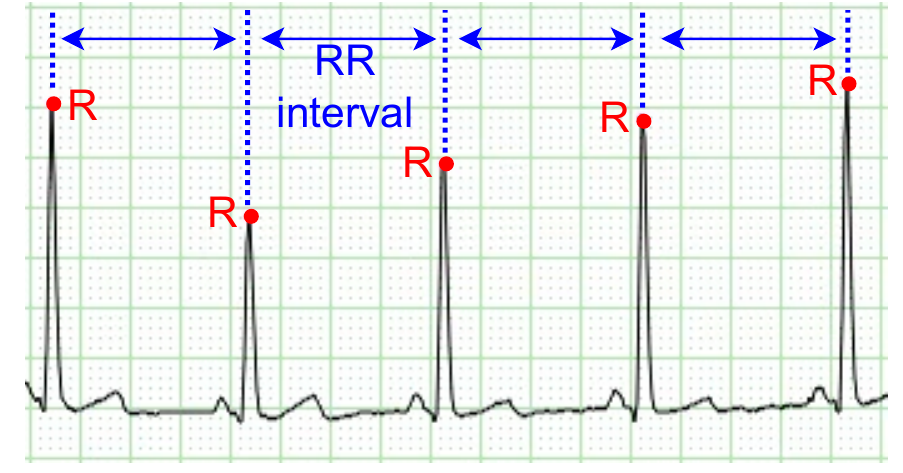}
  \caption{An example of ECG waveform showcasing R peaks (red dots) and RR intervals.}
  \label{fig:ecg_rr_example}
\end{figure}

Human heart rate, when measured as beat-to-beat intervals, is not constant and varies over time~\cite{1965-Schneider-HRV}. This variation, commonly known as Heart Rate Variability (HRV), is an effective indicator of various health and mental problems~\cite{2012-Xhyheri-HRVToday}. The traditional medical device to measure HR and HRV is ECG. ECG records heart activity utilizing electrodes placed at certain skin spots on the human body and produces an electrocardiogram, which is a graph that shows the heart's activity over time.
Electrocardiogram contains the QRS complexes information, which is an important waveform in an electrocardiogram that shows the spread of a stimulus through the ventricles \cite{dohare2014efficient,GOLDBERGER201811}. 
R peaks, which roughly represent heartbeats, can be computed from the QRS complex. The intervals between R peaks are called RR intervals. Figure~\ref{fig:ecg_rr_example} gives an example output from an ECG device that shows the heartbeats and RR intervals. Note that, normal RR intervals are also called NN intervals in the literature~\cite{2012-Xhyheri-HRVToday}.


HRV is defined as the variation of the RR intervals within a time period. HRV is an important health indicator because it represents the adaptive ability of the heart to unpredictable changing circumstances. There is no one standard or best method to calculate HRV \cite{londhe2019heart}. 
In this work, we focus on two commonly used time-domain linear measures for HRV - the Standard Deviation of RR intervals (SDNN) and the Root Mean Square of Successive Differences (RMSSD) \cite{kleiger2005heart}. SDNN is usually recommended for overall HRV estimation and represents both sympathetic and parasympathetic modulation of heart rate, whereas RMSSD is recommended for estimating short-term components of HRV and represents parasympathetic activity~\cite{2012-Xhyheri-HRVToday,1996-Euro-HRV-Circulation}. 

An SDNN or RMSSD is calculated from the RR intervals in a chosen time window, which is usually between 0.5 and 5 minutes~\cite{acharya2006heart}, and a sequence of HRVs over 5 minutes to 24 hours are typically used in medical practice~\cite{1996-Euro-HRV-Circulation}.
The definitions of SDNN and RMSSD are,

\begin{equation}
    \label{eq:sdnn}
    \text{SDNN}=\sqrt{\frac{\sum_{i=1}^{N}(RR_i - \overline{RR})^2}{N}},
\end{equation}

\begin{equation}
    \label{eq:rmssd}
    \text{RMSSD}=\sqrt{\frac{\sum_{i=1}^{N-1}(RR_{i+1} - RR_i)^2}{N-1}},
\end{equation}

where $RR_i$ is the $i^{th}$ RR interval, $\overline{RR}$ is the average, and $N$ is the number of RR intervals within the chosen time window.

Although the ECG produces accurate HRVs, attaching electrodes to the human body can be inconvenient for long-term monitoring~\cite{rajanna2018iot,sutar2013development}. 
Therefore, in this work, we focus on using PPG instead of ECG. 
Nonetheless, we did use ECG to collect reliable HRV readings as the groundtruth to train and evaluate our PPG-based HRV solutions.

\subsection{HRV from PPG}
PPG sensors are popular due to their non-invasive nature. They
are usually attached to human skins at certain locations, such as fingertips, earlobe, and wrist \cite{castaneda2018review}. These sensors utilize infrared light that penetrates the skin to detect changes in the blood circulation -- if there is a change in the blood flow, the intensity of the infrared light also changes~\cite{schafer2013accurate}. Hence, by monitoring the light changes, PPG sensors can detect blood flow changes, which in turn, can be used to infer HR/HRV. 

The main challenge to using PPG for HR/HRV monitoring is light signal noises, especially the motion artifact (MA), which represents the signal noises due to body/hand movements~\cite{fine2021sources}. There are also noises from the environment~\cite{fine2021sources,castaneda2018review} or from the inherent sensor inaccuracy/bias (e.g., sensor sensitivity and calibration issues). Accurate HR/HRV monitoring requires the removal or reduction of these noises~\cite{zhang2014troika,castaneda2018review}.

\subsection{Motivation}\label{sec:motivation}


\begin{figure}
  \centering
  \includegraphics[width=0.8\linewidth]{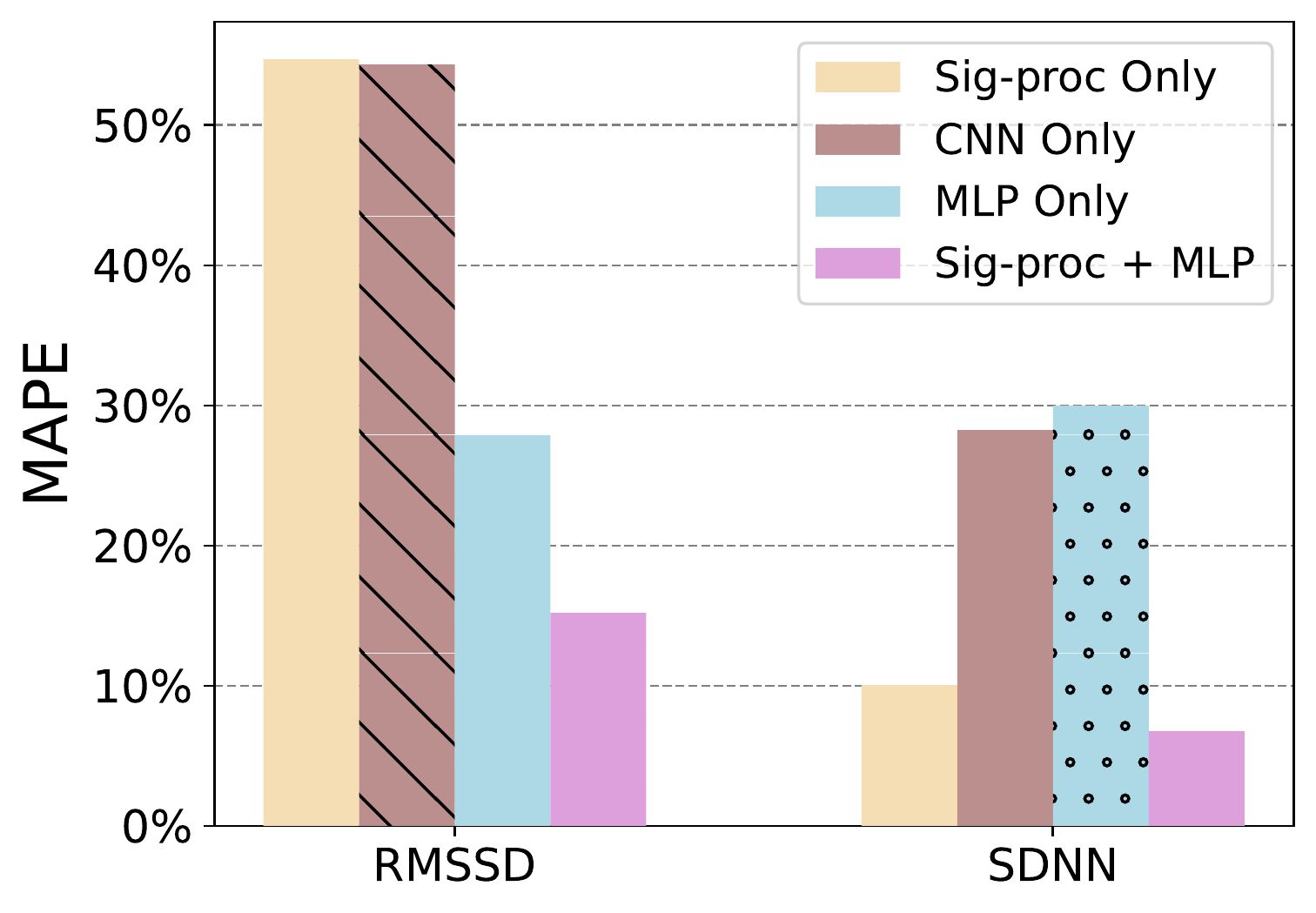}
  \caption{HRV estimation errors of the signal-processing-only \cite{zhang2014troika}, CNN-only \cite{everson2019biotranslator}, MLP-only, and our compound and direct ("Sig-proc+MLP") methods, using the ISPC dataset.}
  \label{fig:motivation_acc}
\end{figure}

\begin{table}
\centering
\begin{tabular}{|c|c|c|c|}
\hline
RR MAPE  & RMSSD MAPE & SDNN MAPE \\ \hline
1\% & 10.24\%  & 4.99\%  \\ \hline
2\% & 27.83\%  & 14.14\% \\ \hline
3\% & 41.55\%  & 23.87\% \\ \hline
4\% & 52.58\%  & 33.19\% \\ \hline
5\% & 61.18\%  & 41.97\% \\ \hline
\end{tabular}
\vspace{2mm}
\caption{RR estimation error amplification after converting to RMSSD and SDNN.}
\vspace{-2mm}
\label{tab:rr_vs_hrv}
\end{table}

\begin{table}[]
\begin{tabular}{|c|c|c|c|}
\hline
HRV   & CNN Only & MLP Only & Sig-proc + MLP \\ \hline
SDNN  & 195KB    & 47.9MB  &  47.9KB   \\ \hline
RMSSD & 195KB    & 122.5MB  &  49.8KB  \\ \hline
\end{tabular}
\vspace{2mm}
\caption{CNN or MLP model sizes of the CNN-only \cite{everson2019biotranslator}, MLP-only, and our compound and direct ("Sig-proc+MLP") methods, using the ISPC dataset. }
\vspace{-2mm}
\label{tab:motivation_size}
\end{table}
Although there have been many studies on applying PPG in HRV inference, these studies were usually limited by their methodology and/or by small datasets. To illustrate these limitations, and as a motivation for this study,  we explored the HRV inference accuracy of a signal-processing-only method~\cite{zhang2014troika} and ML-only method with Convolutional Neural network (CNN) based encoder-decoder~\cite{everson2019biotranslator} using the popular ISPC dataset. 

Figure~\ref{fig:motivation_acc} gives the MAPEs (mean absolute percentage error) of these two methods. For the signal-processing-only method, we reproduced its processing procedures to generate estimations of HRV, which were then compared with the ECG ground truths in the ISPC dataset to calculate the MAPEs. For the CNN-only method, its MAPEs were directly calculated based on results reported in its paper. As Figure~\ref{fig:motivation_acc} shows, both methods have high errors for HRV estimations. Especially for RMSSD, both methods have over 50\% error.  

We observe four reasons that cause these high errors. 
\begin{enumerate}
    \item First, the static signal processing cannot always effectively remove all noises in the PPG signals.
    Therefore, for RMSSD, which measures the short-time components of HRV, the remaining signal noises could significantly degrade the accuracy of the signal-processing-only method. Interestingly, the signal-processing-only method's SDNN estimation accuracy is less affected, as SDNN represents long-term variability and is less sensitive to the remaining signal noises.
    \item Second, the CNN-only method suffers from error amplification. This method does not directly estimate RMSSD or SDNN. Instead, it estimates RR intervals, which are then converted to RMSSD/SDNN using Equations~(\ref{eq:sdnn}) and~(\ref{eq:rmssd}). However, this conversion amplifies the estimation error. Table~\ref{tab:rr_vs_hrv} illustrates this error amplification, where we generated five sets of RR estimations with random errors based on certain average errors (MAPE), converted them into HRV (RMSSD/SDNN), and evaluated the errors of the converted HRV. Table~\ref{tab:rr_vs_hrv} shows that a small 3\% MAPE in RR estimations amplifies to 41.55\%/23.87\% error for RMSSD/SDNN. 
    \item Third, for the CNN-only method, although it can remove most of the noises in theory, the small ISPC dataset does not provide enough data for the CNN model to learn the noise removal completely. The ISPC dataset only has PPG signals from five minutes of monitoring, whereas a single HRV reading requires half to five minutes~\cite{acharya2006heart}. This small dataset further contributes to the CNN model's high errors for both RMSSD and SDNN estimations in Figure~\ref{fig:motivation_acc}. 
    \item Fourth, the model used in the CNN-only method could be too small and not sophisticated enough to process noisy signals.  Table~\ref{tab:motivation_size} shows that this CNN model is only 195KB (42657 trainable parameters). As shown later, inferring HRV with raw noisy PPG signal would require complex neural network models of tens or hundreds of MBs, as the noise removal computation is typically nonlinear and non-polynomial. Large models, however, are unsuitable for small wearable devices with only hundreds of KBs of on-chip memory.
\end{enumerate} 


Based on the above observation, our \textit{\textbf{hypothesis}} is that a \textit{compound and direct method} that combines signal processing and ML, and directly estimates RSMSSD/SDNN, can achieve both high accuracy and small ML model size. To verify this hypothesis, we trained MLP-based models to directly estimate RMSSD and SDNN using the signal-processed ISPC's PPG data. The accuracy and model size of our method are also given in Figure~\ref{fig:motivation_acc} and Table~\ref{tab:motivation_size} under the label "Sig-proc+MLP". As Figure~\ref{fig:motivation_acc} shows, our method had lower error than both the signal-processing-only and ML-only methods. Table~\ref{tab:motivation_size} also shows that this compound and direct method had small model sizes of less than 50KB because many signal noises are already treated by signal processing. 

For the sake of comparison completeness, we also trained MLP models using the original PPG signals from the ISPC dataset to directly estimate RMSSD/SDNN. That is, we also compared an ML-only method with MLP models. These MLP models went through hyperparameter tuning, and the errors of the most-accurate MLP models are reported in Figure~\ref{fig:motivation_acc} under the label "MLP-only", 
which shows that this MLP-only method has lower accuracy than our compound and direct method, mainly due to the small dataset. Moreover, as Table~\ref{tab:motivation_size} shows, the sizes of the MLP-only models are much larger than our compound models, due to the need of relying on pure neural networks to remove all noises.

In summary, the above results show that our hypothesis is likely to be valid, although more data are required to further validate this hypothesis. In the rest of this paper, we will present the details of our methodology for data collection and HRV estimation.

\section{Compound and Direct HRV Estimation}
\label{sec:method}

In this section, we present our signal processing and machine learning combined method for direct HRV estimation.

\subsection{Data Collection}~\label{sec:data_collection}
A single HRV estimation typically requires 30 seconds to 5 minutes of PPG/ECG monitoring~\cite{acharya2006heart}. Moreover, typically, a sequence of five minutes to 24 hours of HRV data is used in medical practice~\cite{2012-Xhyheri-HRVToday}. Therefore, studying HRV estimation requires hours of PPG/ECG monitoring data to provide enough data points. As there lack of such public datasets, our first task in this research was to collect longer traces of PPG and ECG data.

One subject participated in this data collection. The subject had a PPG sensor attached to the fingertip and an ECG monitor attached to the chests at the same time. The PPG data were collected as features/inputs for the HRV inference, whereas the ECG data were used as the labels for model training and as the groundtruth in model testing/evaluation. To reduce the energy consumption caused by the PPG sensors, we collected PPG readings at a frequency of 25Hz, rather than the 125Hz used by the ISPC dataset and other studies~\cite{zhang2015photoplethysmography,bashar2019machine,puranik2019heart,chang2021deepheart}.

The subject conducted three activities during the data collection, including sitting, sleeping, and office work. The office work activity includes actions such as working in front of a computer, walking, and drinking water. For each activity, more than two hours of PPG/ECG data were collected, which gave about 180193 to 180271 PPG readings per activity.
\subsection{Workflow for HRV Inference}

\begin{figure*}
  \centering
  \includegraphics[width=0.8\linewidth]{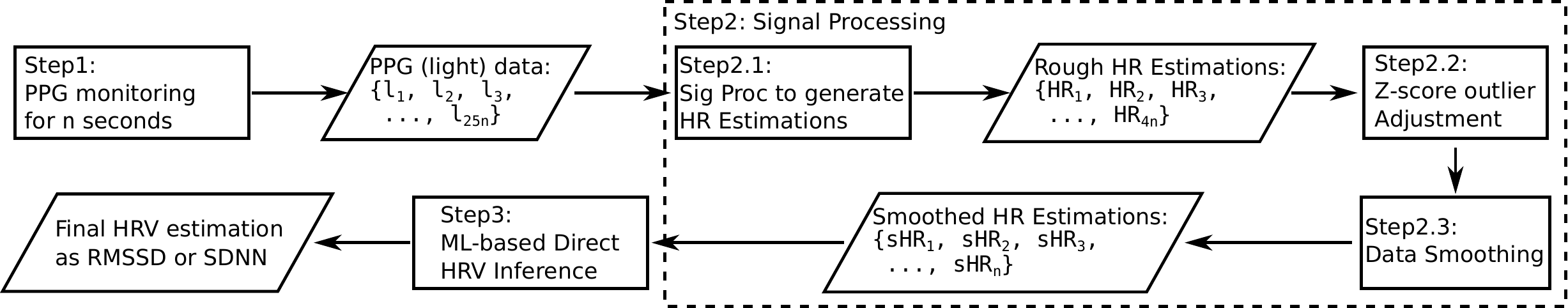}
  \caption{The workflow of our compound and direct HRV inference.}
  \label{fig:stage2}
\end{figure*}

Figure~\ref{fig:stage2} gives the overall workflow of our HRV inference methodology. This workflow includes three major steps, PPG data collection, signal processing, and machine learning-based HRV inference. The following paragraphs provide a detailed description of each step.

\subsubsection{Step 1: PPG Monitoring} 
As our HRV inference is based on PPG sensors, the first step is to collect the PPG light signals reflected from the blood flow. Here, we employed a sampling rate of 25Hz, i.e., 25 signals are collected every second. This sampling rate is lower than the 125Hz used by many prior studies~\cite{zhang2015photoplethysmography,bashar2019machine,puranik2019heart,chang2021deepheart}, and it is employed to reduce the power consumption of sensing, as prior work~\cite{bhowmik2017novel} has shown that high sampling frequency incurs high power usage. Low sampling frequency, however, may negatively affect HRV inference accuracy. As discussed later, we rely on signal processing and machine learning to compensate for this negative accuracy impact. Let $\{l_1, l_2, l_3, \dots, l_{25n}\}$ denotes the sequence of the $25n$ light signals over $n$ seconds. These signals are fed to step 2 for processing.

\subsubsection{Step 2: Signal Processing} 
The primary goal of our signal processing is to remove the noises due to motion artifacts with outlier adjustment and data smoothing. The secondary goal of our signal processing is to generate HR estimations to be used later as features for ML models.

\textbf{Step 2.1: Signal Processing to Generate HR Estimations}. In this step, we convert the PPG light signals collected in Step 1 into four heart rate estimations per second using signal processing. This processing essentially applied a peak detection algorithm to the PPG signals. These peaks can be viewed as "heartbeats", and hence, their counts can be used to estimate HR \cite{ppg_allen2007photoplethysmography}. After this conversion, there are $4n$ HR estimations over $n$ seconds, denoted as $\{HR_1, HR_2, HR_3, \dots, HR_{4n}\}$.

This conversion serves two purposes. First, this conversion simplifies the noise removal (i.e., outlier adjustment and smoothing) conducted later. Motion artifacts typically affect a sequence of PPG light signals, and it can be challenging to distinguish erratic signals from real HR fluctuations when working on raw signals. Converting to HR estimations reduces the number of data points, making it easier for noise removal. Second, these converted HRs are also used as the features of our ML models in Step 3.

\textbf{Step 2.2: Z-score Based Outlier Adjustment}. Many noisy signals can be simply viewed as outliers. Therefore, outlier identification algorithms can be used to remove these signal noises. More specially, we applied the popular Z-score-based outlier filter algorithm~\cite{zscore1_mendenhall2016statistics, zscore2_spiegel2018schaum}.

The Z-score filter identifies the outliers by picking out the data points that deviate the most from the mean. Concretely, consider a large sequence of HR estimations, $\{HR_1, HR_2, \dots\}$, with mean $\mu$ and standard deviation $\delta$. If the difference between a data point $HR_i$ and $\mu$ ( i.e., $|HR_i-\mu|$) is larger than a threshold, then $HR_i$ can be viewed as an outlier. This threshold is typically defined based on the standard deviation $\delta$. Particularly, the threshold is defined as $z\_score\times\delta$. In this work, we used a z\_score of $3$, following common practices~\cite{zscore3_zhang2011illustration, zscore4_vysochanskij1980justification}. That is, if the difference between HR estimation $HR_i$ and the mean $\mu$ is larger than $3\delta$, then $HR_i$ is deemed as an outlier.

 We do not remove outliers because HR estimations are time series, and removing estimations would create "holes" within the time series. Consequently, after the outliers are identified, their values are just adjusted to be the average of their neighboring estimations. That is, given an outlier $HR_i$, its value is adjusted to be $\frac{HR_{i-1} + HR_{i+1}}{2}$.
 
 \textbf{Step 2.3: Data Smoothing}. The above outlier adjustment only identifies data points with large noises, i.e., data points deviate greatly from the overall mean. However, there could still be HR estimations that deviate significantly from local HR averages. These deviating HR estimations usually manifest themselves as abnormal local peaks/valleys. These local peaks/valleys are usually caused by signal noises, because normally, a person's heart rate does not increase (or drop) abruptly and drops (or increases) back within one second. 
 
 To remove these local noises, we apply moving average data smoothing. More specifically, this data smoothing converts the four rough HRs within a second into one HR estimation per second by computing their averages to reduce the impact of the abnormal local peaks and valleys.
 After the smoothing, there are $n$ smoothed HR estimations over $n$ seconds, denoted by $\{sHR_1, sHR_2, sHR_3, \dots\, sHR_n\}$. 
 

\subsubsection{Step 3: ML-based HRV Inference}\label{sec:ml_hrv_models} 
The last step of our methodology employs ML to infer HRV. As stated previously, the ML models are trained to serve two purposes simultaneously. First, they are trained to further remove/reduce the noises that cannot be filtered by signal processing. These additional noises may include long motion artifacts that affect several seconds of PPG signals, the sensor bias (e.g., sensor sensitivity and calibration issues), and
the errors due to our low sampling frequency. Second, with further reduced noises, these ML models estimates the final HRV. These models directly infer SDNN/RMSSD, instead of the RR intervals.

The main input features to the ML models are the smoothed HR estimations over $n$ seconds from Step 2, i.e., the vector $\{sHR_1, sHR_2, sHR_3, \dots\, sHR_n\}$. These smoothed HR estimations are also used to derive a rough HRV (SDNN or RMSSD) estimation, denoted by $rHRV$, using the Equations~(\ref{eq:sdnn}) and~(\ref{eq:rmssd}). This rough HRV is used as a constructed feature for our ML models. As shown with Equations~(\ref{eq:sdnn}) and~(\ref{eq:rmssd}), computing HRV requires exponentiation and square root operations, which may require large ML models to simulate. Therefore, employing a computed rough HRV estimation as a feature can potentially reduce the trained model size. 

In summary, the features of our ML models are the vector,
$\{sHR_1, sHR_2, sHR_3, \dots\, sHR_n, rHRV\}$. The output of our models is the HRV estimation for the past $n$ seconds in SDNN or RMSSD. Note that, $n$ represents the number of seconds of PPG monitoring, which is a tuneable parameter depending on the HRV use case. In the experimental evaluation (Section~\ref{sec:Evaluation}), we evaluated different values for $n$.

\subsection{Model Training}
During our PPG data collection (described in Section~\ref{sec:data_collection}), the subject also had an ECG attached to collect groundtruth HRV readings. The groundtruth HRVs are used both as the labels in the training data and validation/testing data. 

We partitioned the collected data into training and testing datasets with a split of 80\% and 20\%. All models were also optimized with hyperparameter tuning~\cite{bergstra2012random} to find the model with the best accuracy. The specific hyperparameters for each type of model are given in Section~\ref{sec:hyperparams}. Note that, we limited hyperparameter search space to avoid generating models that are too large for small embedded devices.

\section{Experimental Evaluation}
\label{sec:Evaluation}

This section presents the evaluation results, focusing on the HRV inference accuracy, ML model sizes, and inference time.

\subsection{Experiment Setup}



\subsubsection{Hardware used in Data Collection and Inference}


Our dataset contains one channel of PPG data. 
The subject simultaneously wore the PPG sensor on the fingertip 
and ECG electrodes. The PPG sensor 
is connected to a Raspberry Pi 4B through an i2C bus to record the data.
The hardware components are: 1) a Raspberry Pi 4B with 4 GB RAM; 2) a Maxim MAXREFDES117 HR monitor with a MAX30102 PPG sensor; and 3) a TLC5007 Dynamic ECG. 
PPG data and ECG data are synchronized according to their timestamps.
The Raspberry Pi is also used to measure the inference latency of the signal processing and ML models.







\subsubsection{ML Models and Hyperparamter Configurations}~\label{sec:hyperparams}
DT, RF, KNN, and SVM models were implemented using Scikit-learn~\cite{scikit-learn} and MLP was implemented using Keras.
ML models were saved with $joblib$.
We used the random search function $RandomizedSearchCV$ in Scikit-learn and $RandomSearch$ in Keras-tuner to tune the model hyperparameters. 
The hyperparameters are: 
1) For DT, the maximum depth of the tree ranges from 3 to 20; 
2) For RF, the number of trees is between 2 to 128, and the maximum tree depth is between 3 to 20;
3) For KNN, the number of neighbors is between 2 to 30, and the distance can be Manhattan or Euclidean; 
4) For SVM, the kernel may be among RBF, sigmoid and polynomial, and regularization (i.e., $C$) may be between 0.00001 to 10; 
5) For MLP, there are 1 to 5 hidden layers, each with 1 to 100 neurons, and the activation function can be $relu$ or $tanh$.

\subsubsection{Metrics}~\label{sec:metrics}
For HRV estimation accuracy evaluation, we used the metric MAPE (Mean Absolute Percentage Error) between the HRV estimations and groundtruths. 
Given $m$ HRV estimations, the definition of MAPE is,
\begin{equation}
    MAPE = \frac{100\%}{m}\sum^{m}_{i=1}\bigg| \frac{HRV_{est,i}-HRV_{true,i}}{HRV_{true,i}}\bigg|.
\end{equation}

For model size evaluation, we report the sizes in KB (kilobytes) or MB (megabytes).
For inference time evaluation, we report the average time it takes to make an inference in microseconds ($\mu$s) or milliseconds (ms). 

\subsection{Accuracy Evaluation}

\begin{figure*}
  \subfloat[Office Work.]
  {\includegraphics[width=0.33\textwidth]{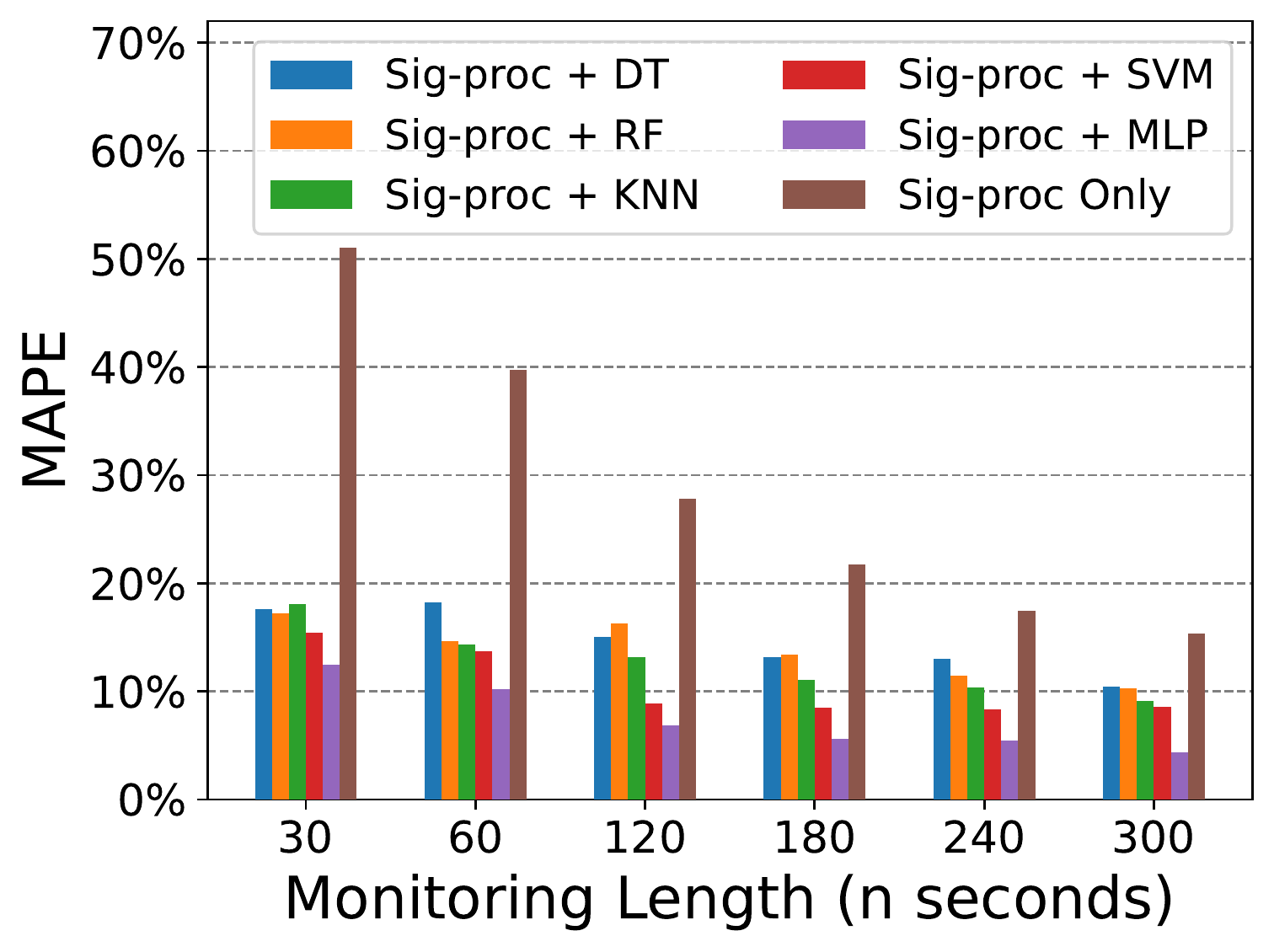}\label{fig:mape_rmssd_daily}}
  \hfill
  \subfloat[Sleep.]
  {\includegraphics[width=0.33\textwidth]{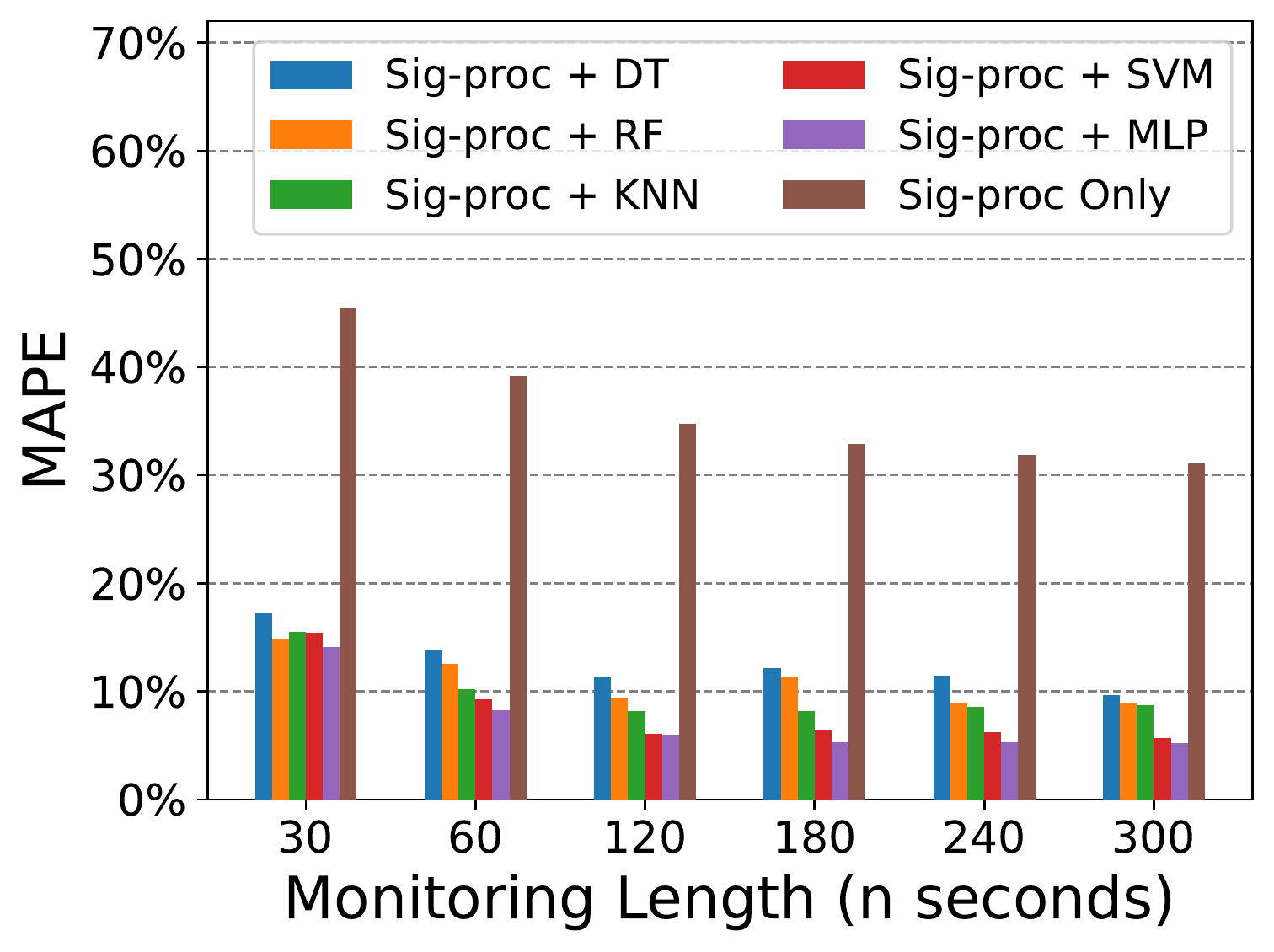}\label{fig:mape_rmssd_sleep}}
  \hfill
  \subfloat[Sit.]
  {\includegraphics[width=0.33\textwidth]{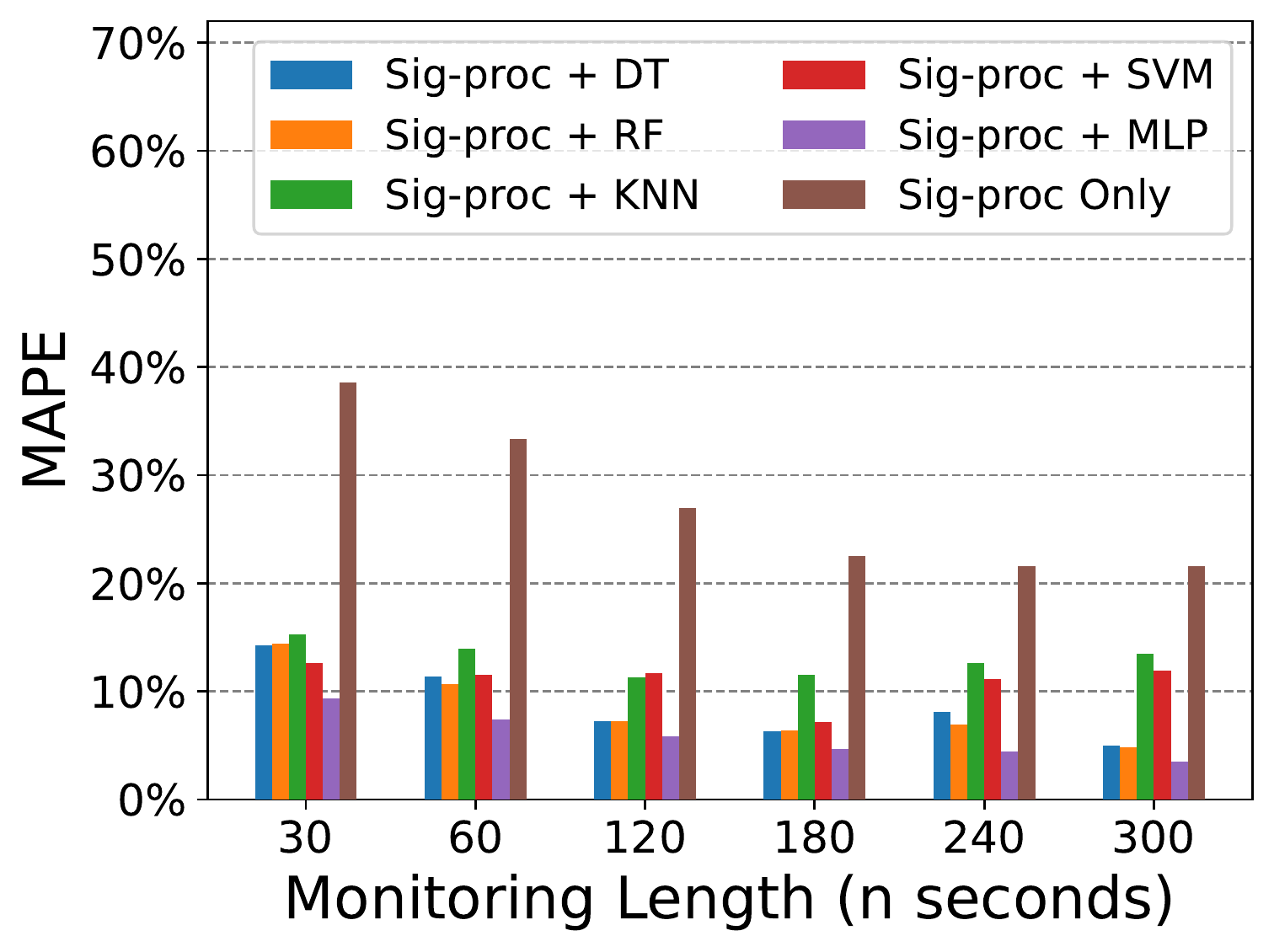}\label{fig:mape_rmssd_sit}}
  \caption{MAPEs for HRV/RMSSD estimations for different activities.}
  \label{fig:mape_rmssd}
\end{figure*}

\begin{figure*}
  \subfloat[Office Work.]
  {\includegraphics[width=0.33\textwidth]{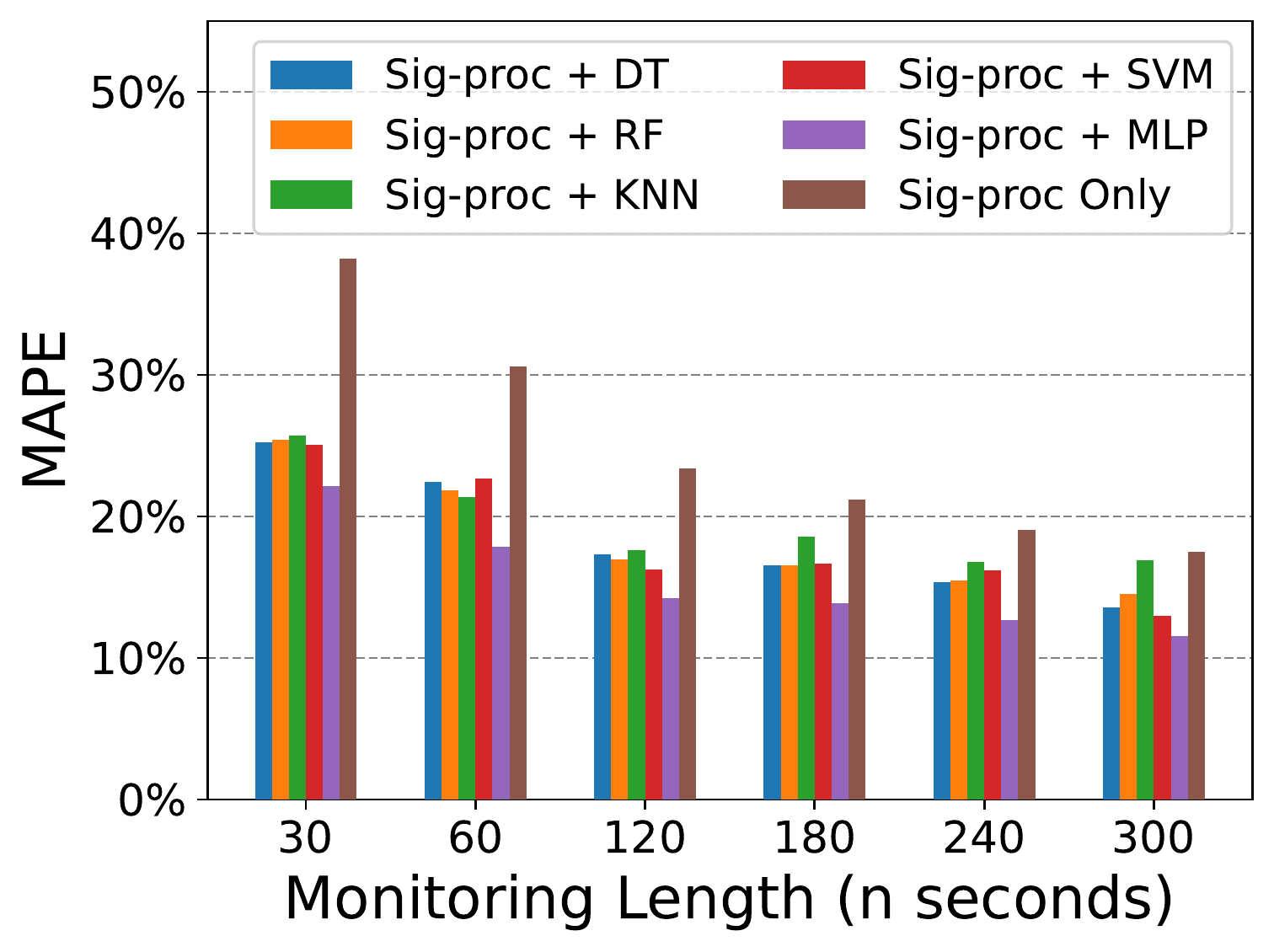}\label{fig:mape_sdnn_daily}}
  \hfill
  \subfloat[Sleep.]
  {\includegraphics[width=0.33\textwidth]{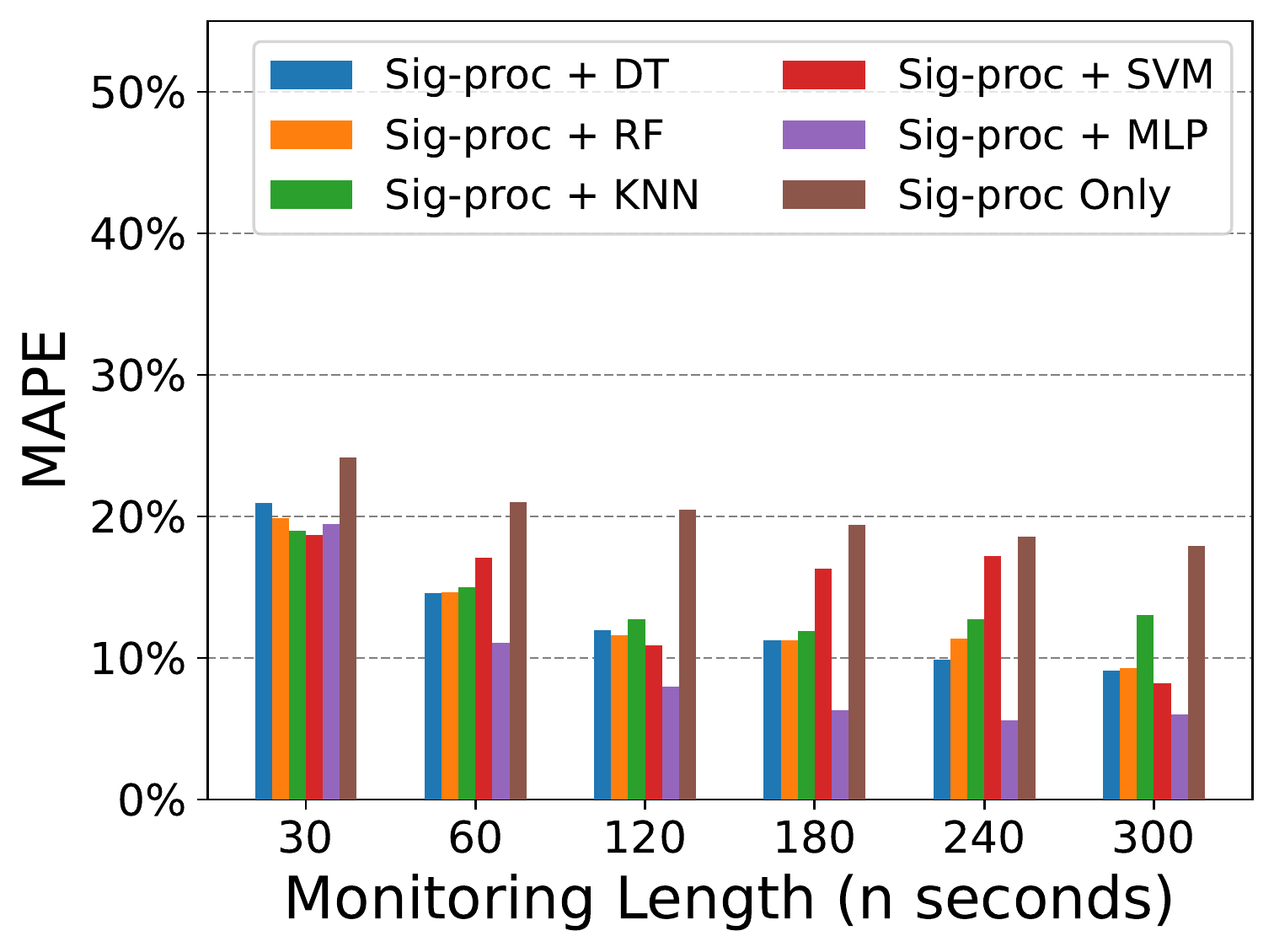}\label{fig:mape_sdnn_sleep}}
  \hfill
  \subfloat[Sit.]
  {\includegraphics[width=0.33\textwidth]{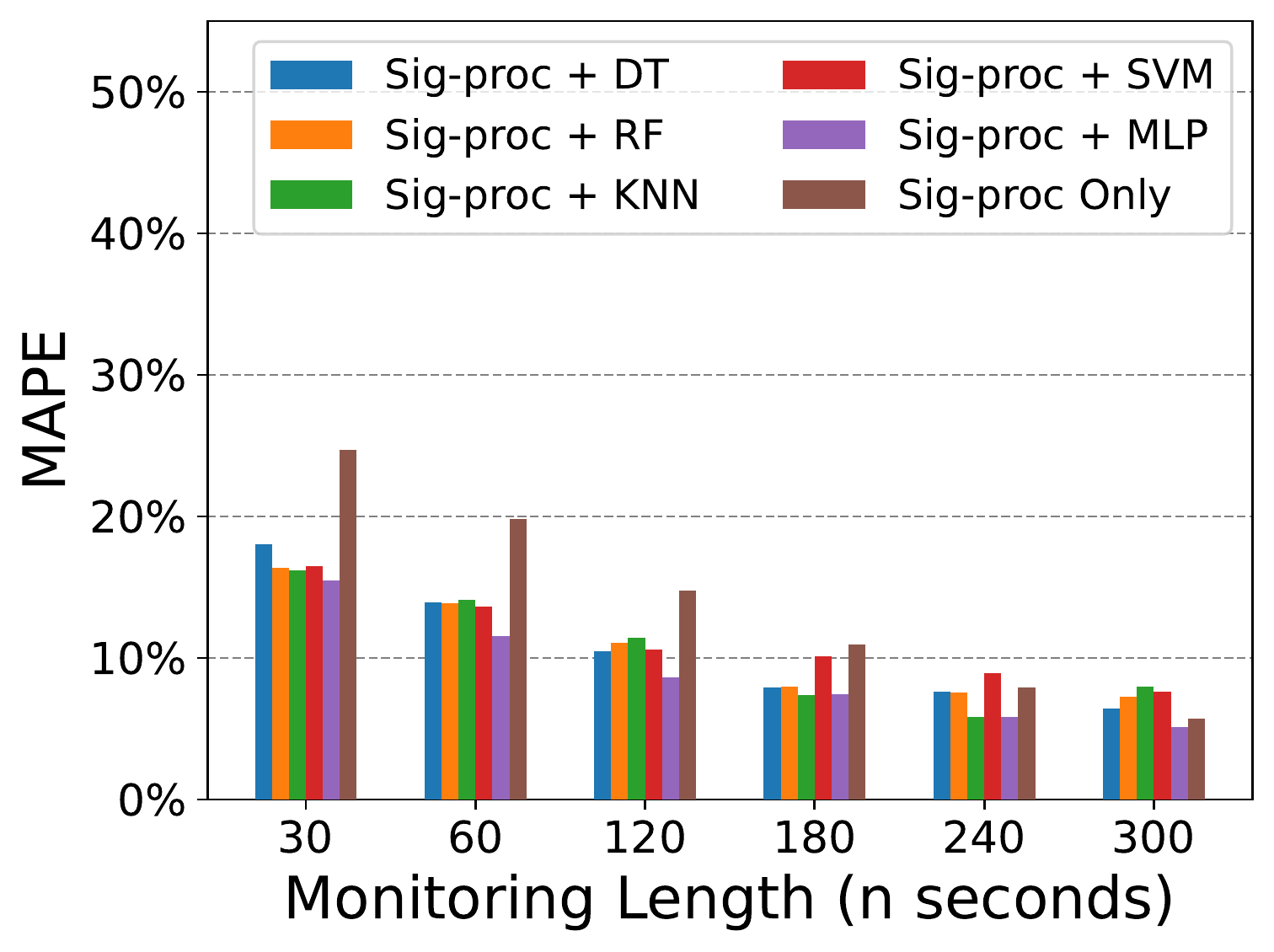}\label{fig:mape_sdnn_sit}}
  \caption{MAPEs for HRV/SDNN estimations for different activities.}
  \label{fig:mape_sdnn}
\end{figure*}



\begin{figure}
  \centering
  \includegraphics[width=0.8\linewidth]{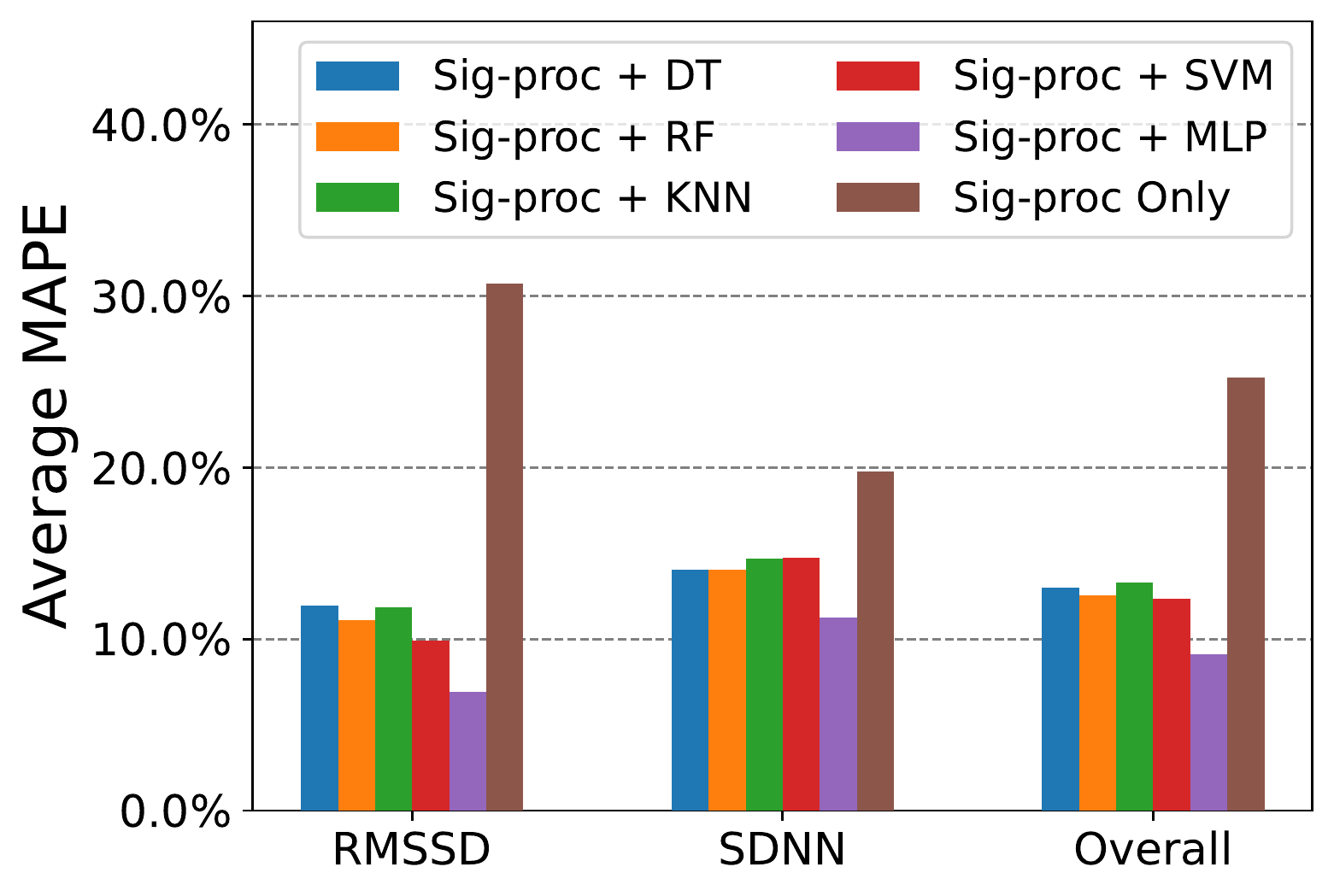}
  \caption{Average MAPEs across all activities and all monitoring lengths.}
  \label{fig:average_mapes}
\end{figure}

Figure~\ref{fig:mape_rmssd} and Figure~\ref{fig:mape_sdnn} give the MAPE of our compound and direct methods, using different types of ML algorithms at different monitoring lengths (i.e., the $n$ seconds in Section~\ref{sec:ml_hrv_models}). The figures also give the errors of the signal-processing-only method ("Sig-proc Only"). The following paragraphs discuss these accuracy results in detail.

\subsubsection{Overall Accuracy of Our Method}
Figure~\ref{fig:mape_rmssd} and Figure~\ref{fig:mape_sdnn} show that our compound and direct prediction methods usually had errors between $3.5\%$ to $25.7\%$. 
The highest error was $25.7\%$, which was for the KNN model for SDNN estimation under the office work scenario with 30 seconds monitoring length (Figure~\ref{fig:mape_sdnn_daily}). The lowest error was only $3.5\%$, for the MLP model for RMSSD inference under sit scenario with 300 seconds monitoring length (Figure~\ref{fig:mape_rmssd_sit}). 

For the majority of the models, the errors of our method were less than 20\%. Figure~\ref{fig:average_mapes} also shows that the overall average MAPEs for different ML models (overall activities for both RMSSD and SDNN) are all less than 13.2\%. MLP models have the lowest overall average MAPEs of only 9.1\%. These results show that our compound and direct method has high accuracy for HRV estimation.

\subsubsection{Comparison with Signal Processing Only}
Figure~\ref{fig:mape_rmssd} and Figure~\ref{fig:mape_sdnn} also show that our method was usually more accurate than the signal-processing-only method ("Sig-proc Only"). In the case of RMSSD estimation (Figure~\ref{fig:mape_rmssd}), the signal-processing-only method usually had errors above 20\%, whereas our method's errors are usually less than 20\%.  

In the SDNN estimations (Figure~\ref{fig:mape_sdnn}), 
the signal-processing-only method had lower errors than their RMSSD estimations, because SDNN measures long-term HRV and is less sensitive to signal noises. We had a similar observation for the ISPC dataset as discussed in Section~\ref{sec:motivation}. Nonetheless, our method usually still had lower errors than the signal-processing-only method for SDNN estimations. 

The only exceptions were for SDNN estimations in sit scenario with 240/300 seconds monitoring lengths (Figure~\ref{fig:mape_sdnn_sit}, where our method with some ML models (e.g., SVM) had higher errors than the signal-processing-only method. However, the error difference was small -- only 2.2\% at most. Moreover, our method with the MLP model was still more accurate than signal-processing-only in these cases.

Overall, Figure~\ref{fig:average_mapes} shows that the signal-processing-only method has an overall average MAPE of 25\%, which is higher than the overall average MAPEs (about 12\%) of our compound and direct method using any ML model. 

\subsubsection{Traces of HRV Estimation}

\begin{figure*}
  \centering
  \includegraphics[width=1\linewidth]{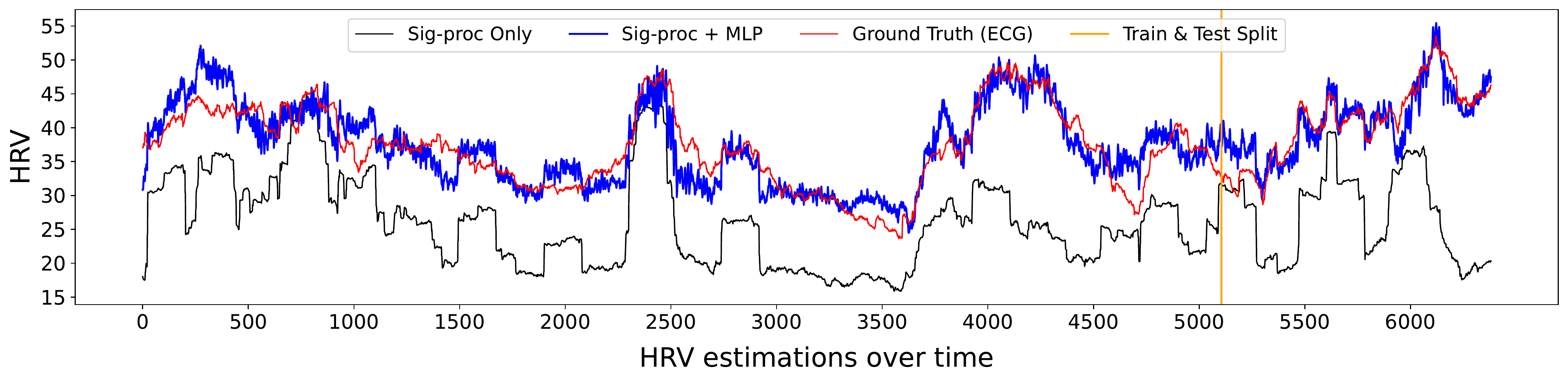}
  \caption{RMSSD estimation trace (sleep activity, monitoring length of 180sec). The yellow line separates training and test data.} 
  \label{fig:trace_rmssd}
\end{figure*}

\begin{figure*}
  \centering
  \includegraphics[width=1\linewidth]{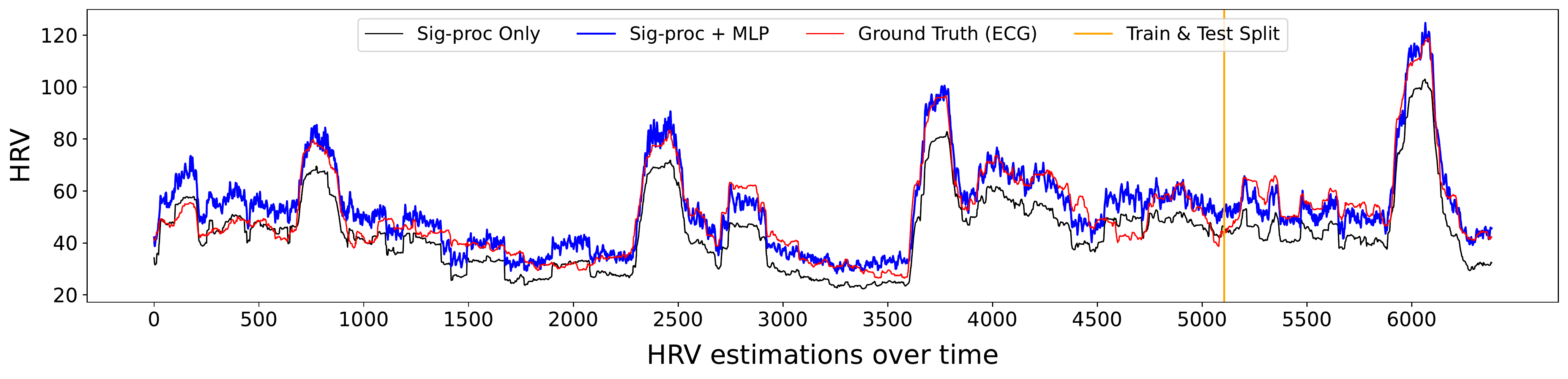}
  \caption{SDNN estimation trace (sleep activity, monitoring length of 180sec). The yellow line separates training and test data.} 
  \label{fig:trace_sdnn}
\end{figure*}

Figure~\ref{fig:trace_rmssd} and Figure~\ref{fig:trace_sdnn} give the traces of the RMSSD HRVs from the groundtruth (ECG), signal-processing-only method, and our method with MLP model for the sleep activity data. These traces illustrate that our MLP models reduce at least two types of noises after signal processing. The first type of noise is mostly errors due to either sensor bias or low sampling frequency. For example, for the 5600'th to 5750'th HRVs in Figure~\ref{fig:trace_rmssd}, signal-processing HRVs had similar fluctuation trends as the ECG HRVs, but they deviate by roughly 10. The MLP model, however, was trained to correct this "deviation" and produced more accurate HRVs. This same issue can also be observed for the SDNN estimations in Figure~\ref{fig:trace_sdnn}. The second type of noise is usually from motion artifact noises that affect a longer period. For example, for the 5900'th to 6200'th HRVs in Figure~\ref{fig:trace_rmssd}, signal-processing HRVs were flat then dropped, whereas the ECG HRVs were sharply increasing. These longer errors were also detected and corrected by our MLP model to produce HRVs increasing from 38 to 50, similar to the ECG.

The same conclusion can be drawn from the traces for other models, activities, and monitoring lengths. However, due to space limitations, these traces are omitted.

\subsubsection{Impact of Type of Activity and HRV Metric}
Across the results for the three activities in Figure~\ref{fig:mape_rmssd} and Figure~\ref{fig:mape_sdnn}, HRV estimations for sit had the lowest MAPEs. These low errors were because when the subjects sat, they had little movement, hence, lower motion artifacts. 
In this work, we built one ML model for each activity. However, if only one ML model is built to cover all activities, then these MAPE differences suggest that more sensors/features (e.g., accelerometer or gyroscope) may be needed to conduct noise correction differently for different activities.

Moreover, 
Figure~\ref{fig:average_mapes} also show that our method's average MAPEs only differ slightly for RMSSD and SDNN estimations, indicating that our method works for both short-term or long-term HRV monitoring. However, the signal-processing-only method had considerably lower errors for SDNN (long-term) estimations than RMSSD (short-term) estimations.

\subsubsection{Accuracy Impact of Model Types}
Figure~\ref{fig:average_mapes} shows that MLP models were generally more accurate than other ML models in our method, with average MAPEs for RMSSD, SDNN, and "overall" being only 6.9\%, 11.3\%, and 9.1\%, respectively. Nonetheless, the differences in average MAPEs among model types in Figure~\ref{fig:average_mapes} are less than 4\%. 
These similar MAPEs match the recently proposed "Rashomon" theory~\cite{Rashomon-ML}, which states that there could be multiple ML models having similar accuracy for the same dataset. A group of models with similar accuracy implies that it is beneficial to conduct ML model exploration to search for models that fit certain non-functional requirements, such as low model sizes and fast inference time in the embedded applications. 


\begin{figure*}
  \subfloat[Office Work.]
  {\includegraphics[width=0.33\textwidth]{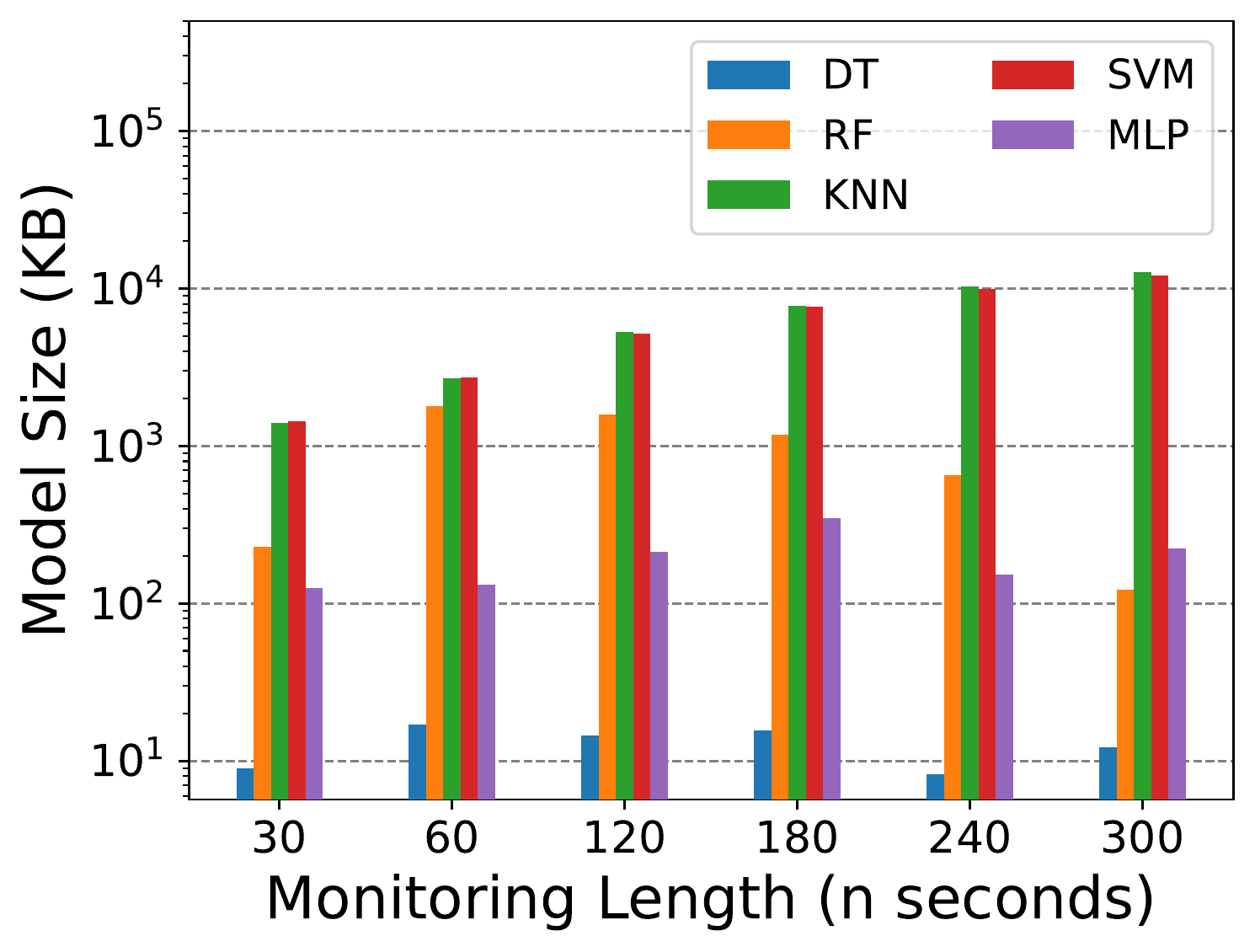}\label{fig:size_rmssd_daily}}
  \hfill
  \subfloat[Sleep.]
  {\includegraphics[width=0.33\textwidth]{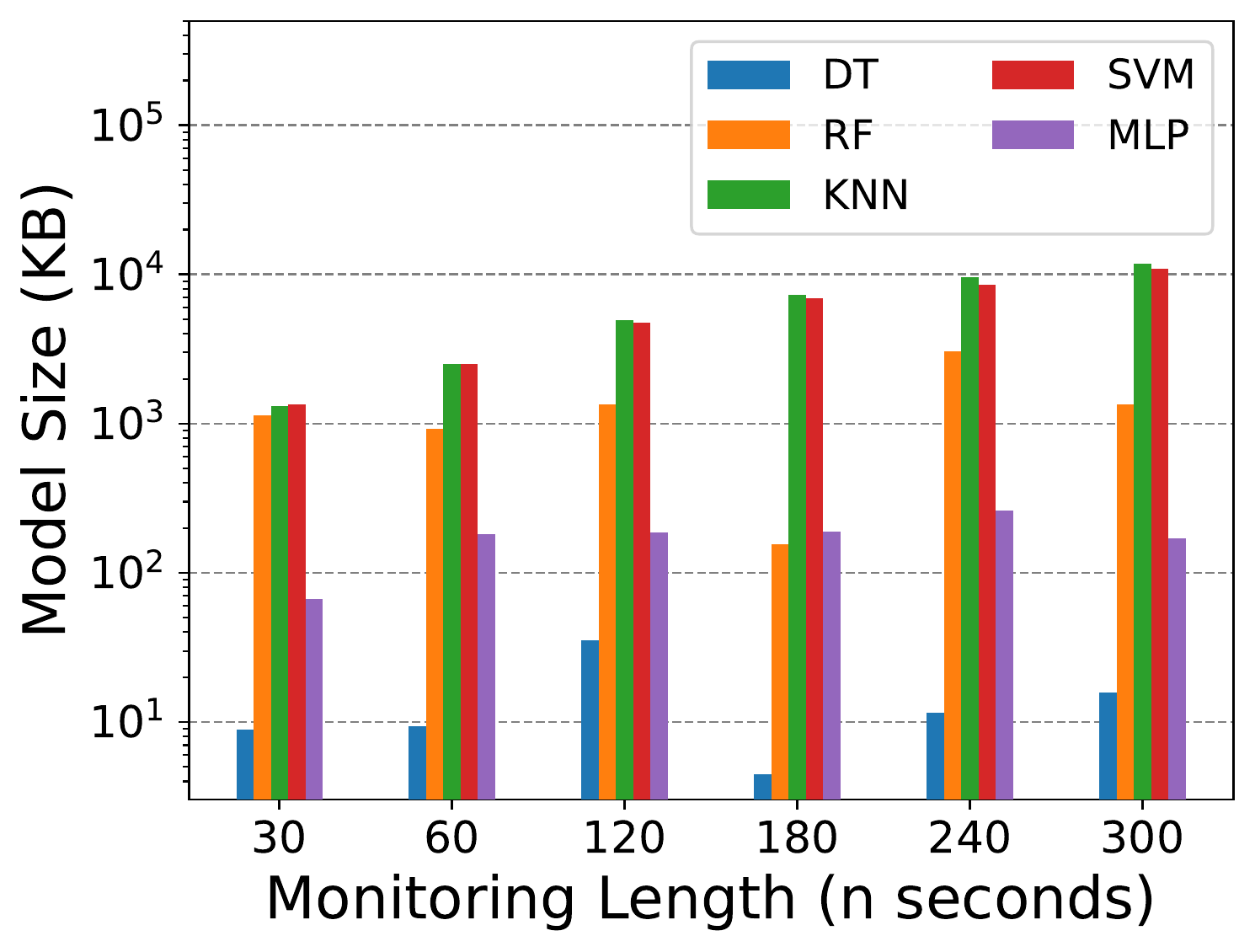}\label{fig:size_rmssd_sleep}}
  \hfill
  \subfloat[Sit.]
  {\includegraphics[width=0.33\textwidth]{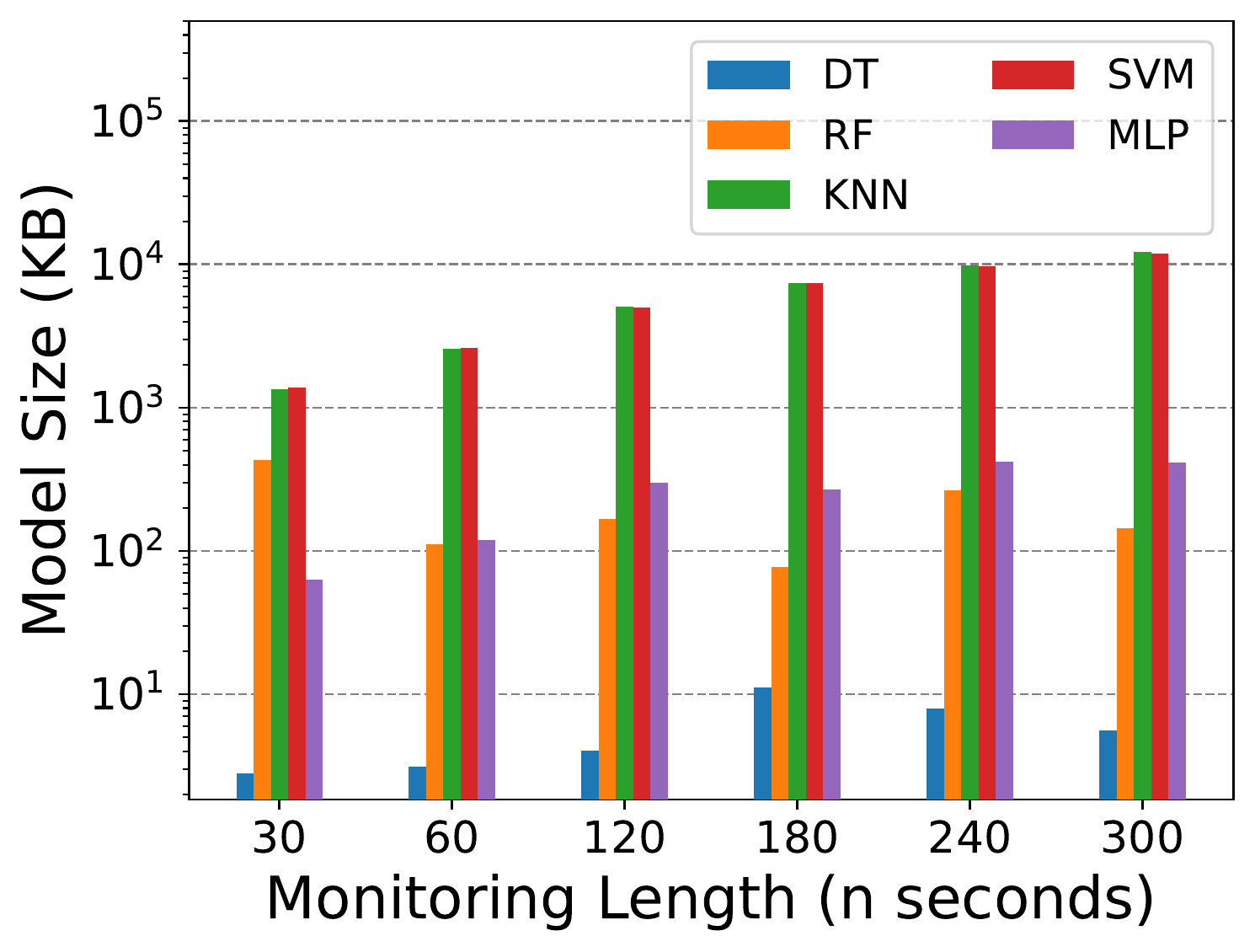}\label{fig:size_rmssd_sit}}
  \caption{Sizes of HRV/RMSSD models for different activities.}
  \label{fig:size_rmssd}
\end{figure*}

\begin{figure*}
  \subfloat[Office Work.]
  {\includegraphics[width=0.33\textwidth]{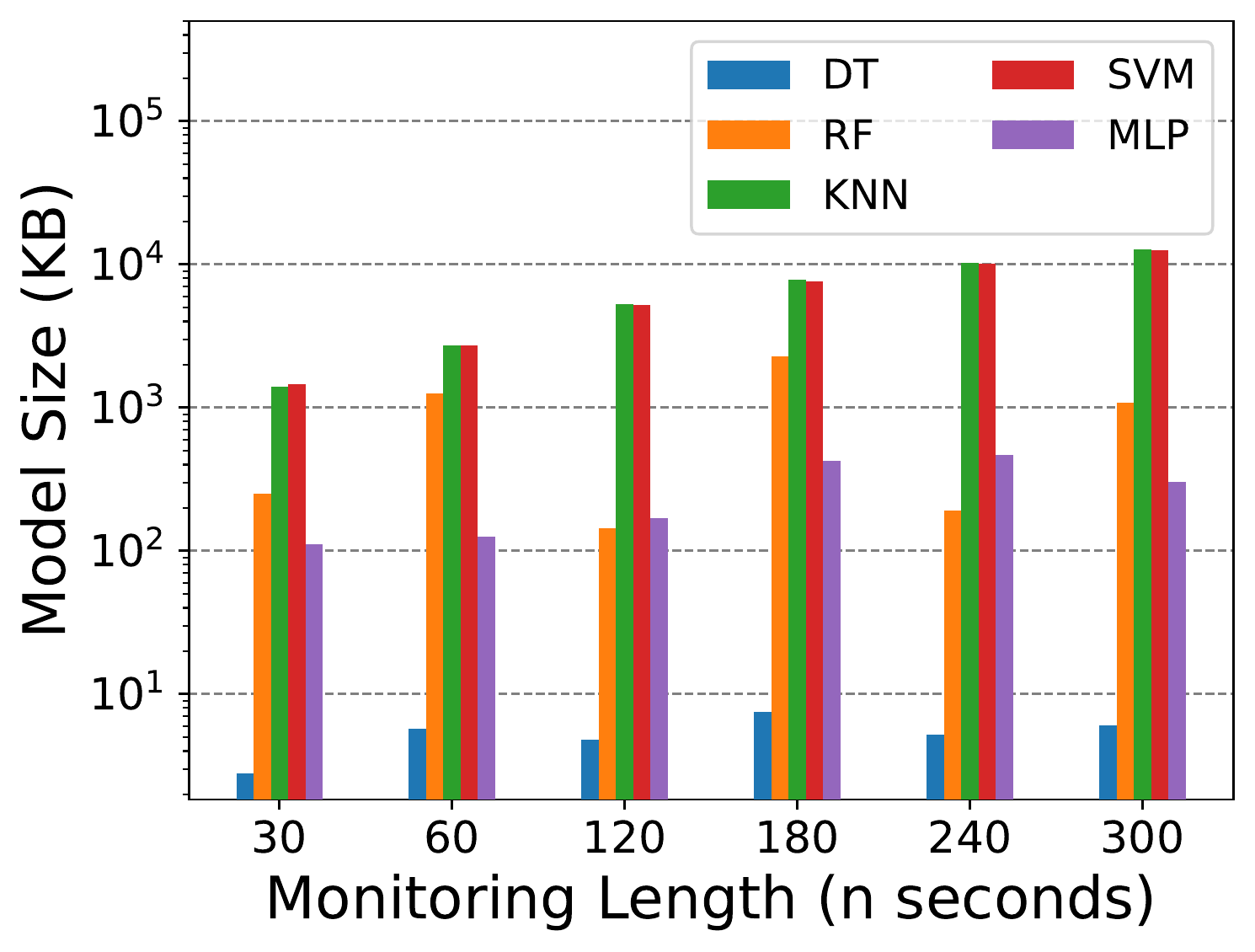}\label{fig:size_sdnn_daily}}
  \hfill
  \subfloat[Sleep.]
  {\includegraphics[width=0.33\textwidth]{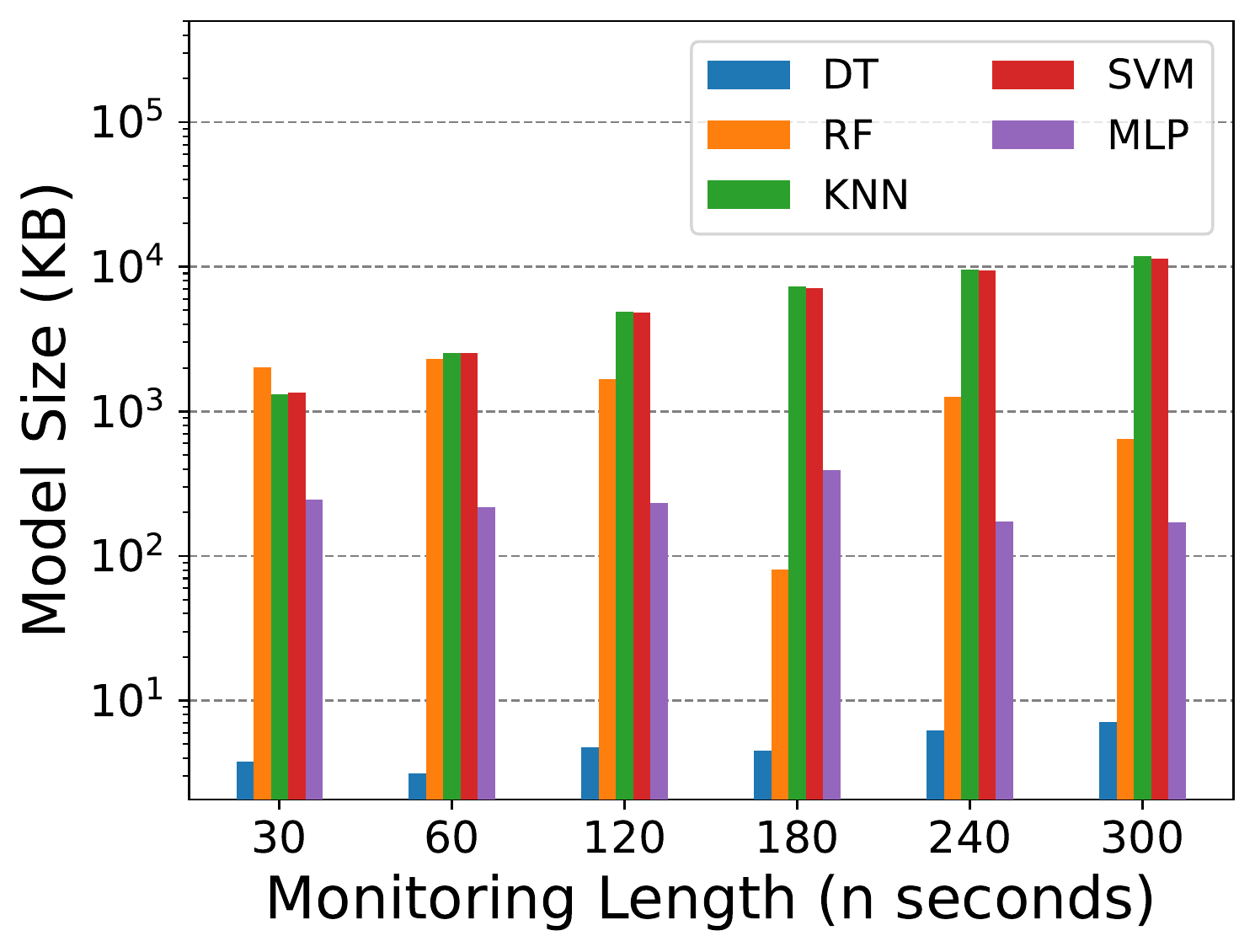}\label{fig:size_sdnn_sleep}}
  \hfill
  \subfloat[Sit.]
  {\includegraphics[width=0.33\textwidth]{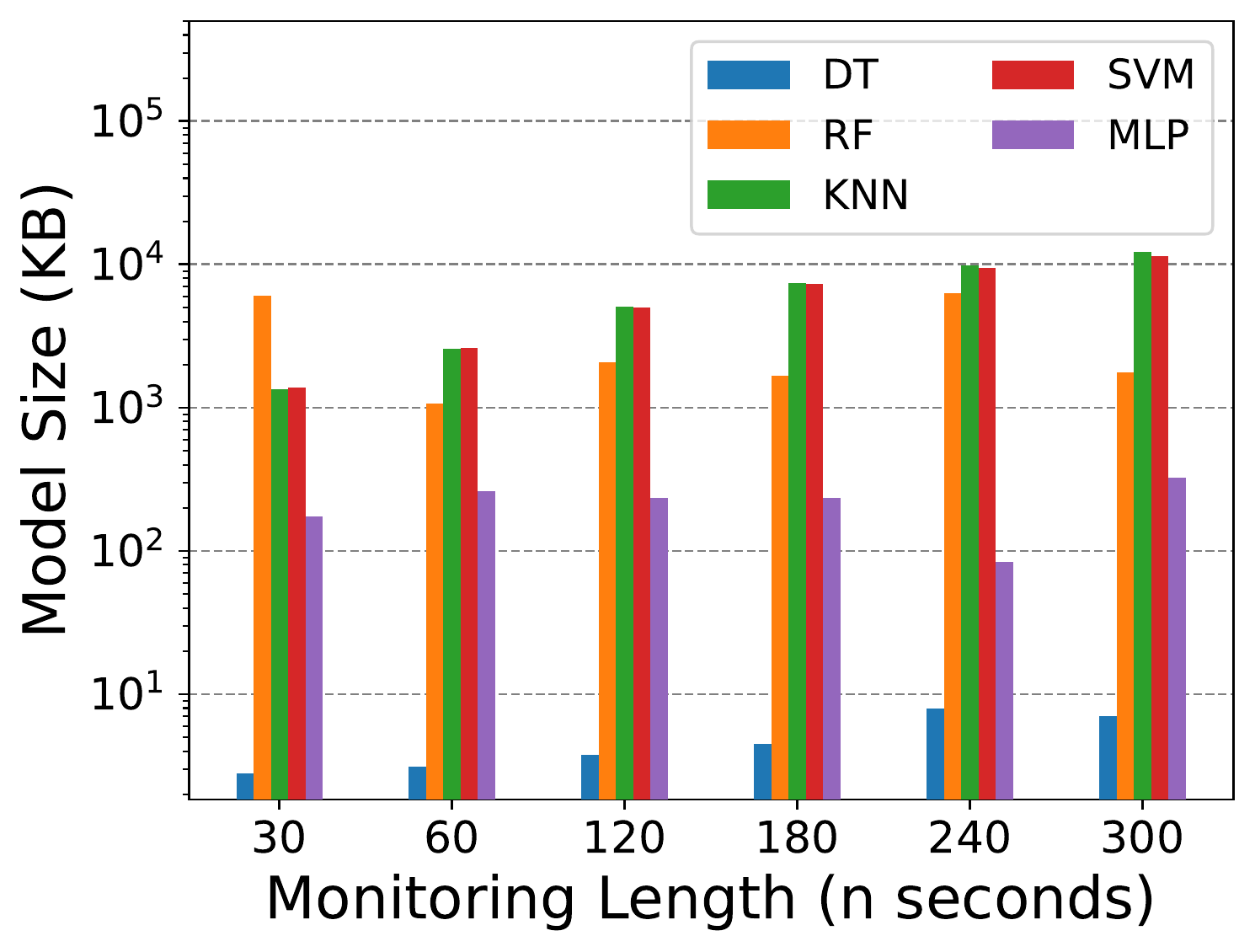}\label{fig:size_sdnn_sit}}
  \caption{Sizes of HRV/SDNN models for different activities.}
  \label{fig:size_sdnn}
\end{figure*}
\subsubsection{Accuracy Impact of Monitoring Length}
Figure~\ref{fig:mape_rmssd} and Figure~\ref{fig:mape_sdnn} also show that another factor that impacts the accuracy of our method is the monitoring length $n$. That is, the longer the monitoring length, the lower the HRV estimation error. And the HRV estimations for the monitoring length at 300 seconds usually had the lowest errors. An HRV estimation for longer monitoring is usually less susceptible to a few noisy PPG signals, and therefore, it tends to have lower errors.

Note that, the required HRV monitoring length depends on the use case~\cite{2012-Xhyheri-HRVToday}. For example, 300-second monitoring is applied for caring for chronic renal failure and diabetes~\cite{acharya2006heart}. The fact that errors vary with monitoring lengths also suggests the applicability of PPG-based HRV monitoring to medical use needs to be evaluated case by case.

\subsubsection{Comparison with the ML-only Method}
We also experimented with the ML-only method to directly infer HRV using the original PPG data collected from the sensors. However, we were not able to obtain any ML-only HRV models with good accuracy, or even meaningful estimations. These models typically have accuracy similar to, sometimes even worse than, the signal-processing-only method. For the case of MLP, the HRV estimations produced by each MLP model are mostly the same value, making the estimations practically useless. These MLP models also have average sizes of 6.7MB and maximum size of 26.6MB, larger than the MLP models in our compound method.

We believe the main cause of the low accuracy was the small hyperparameter search space. Recall that we limited the hyperparameter search space to limit the ML model sizes, which also limits the model complexity. However, it typically needs very complex ML models to infer HRV using only the original PPG signals. Because of the poor results of the ML-only HRV models, we did not include them in the paper.


\begin{figure*}
  \subfloat[Office Work.]
  {\includegraphics[width=0.33\textwidth]{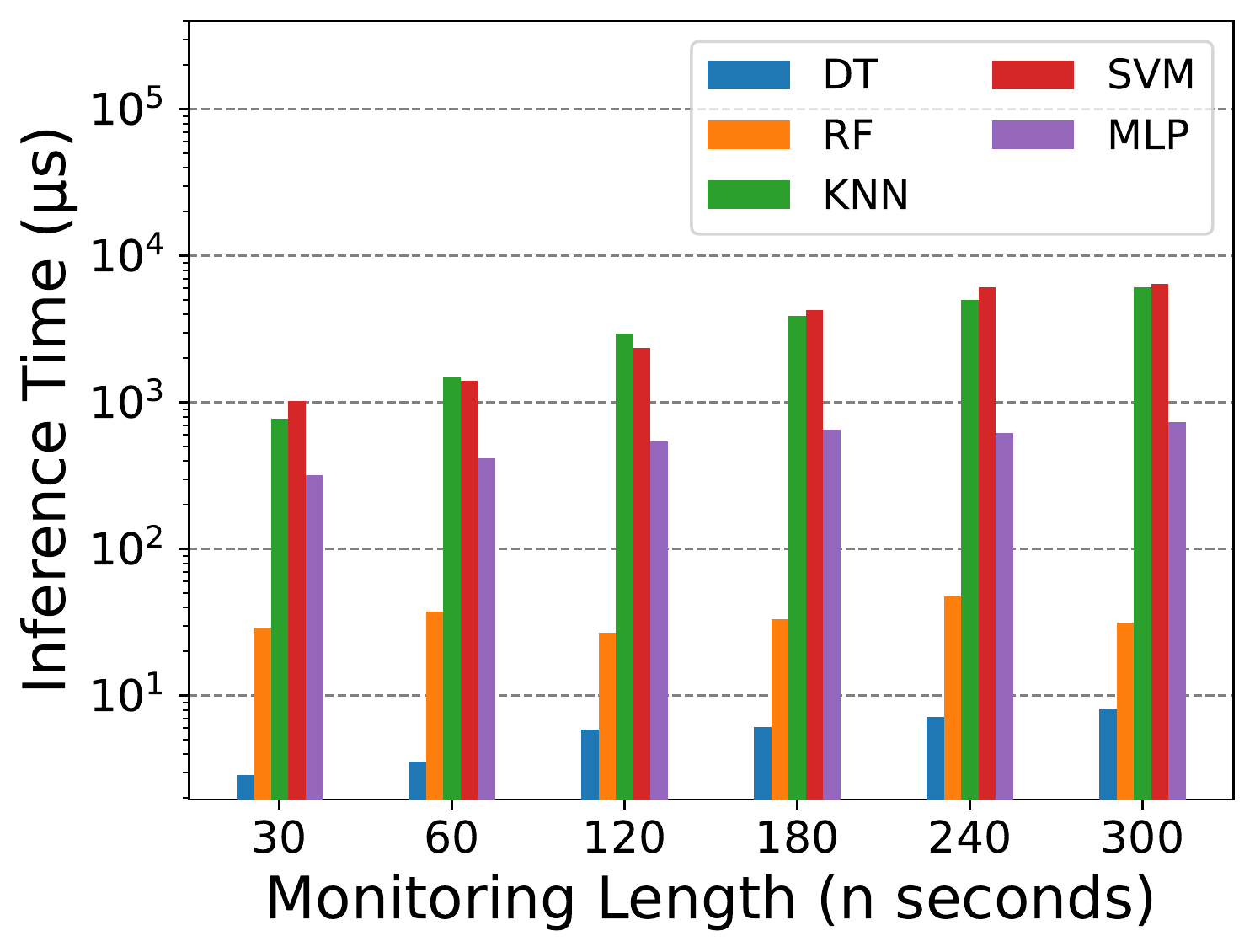}\label{fig:time_rmssd_daily}}
  \hfill
  \subfloat[Sleep.]
  {\includegraphics[width=0.33\textwidth]{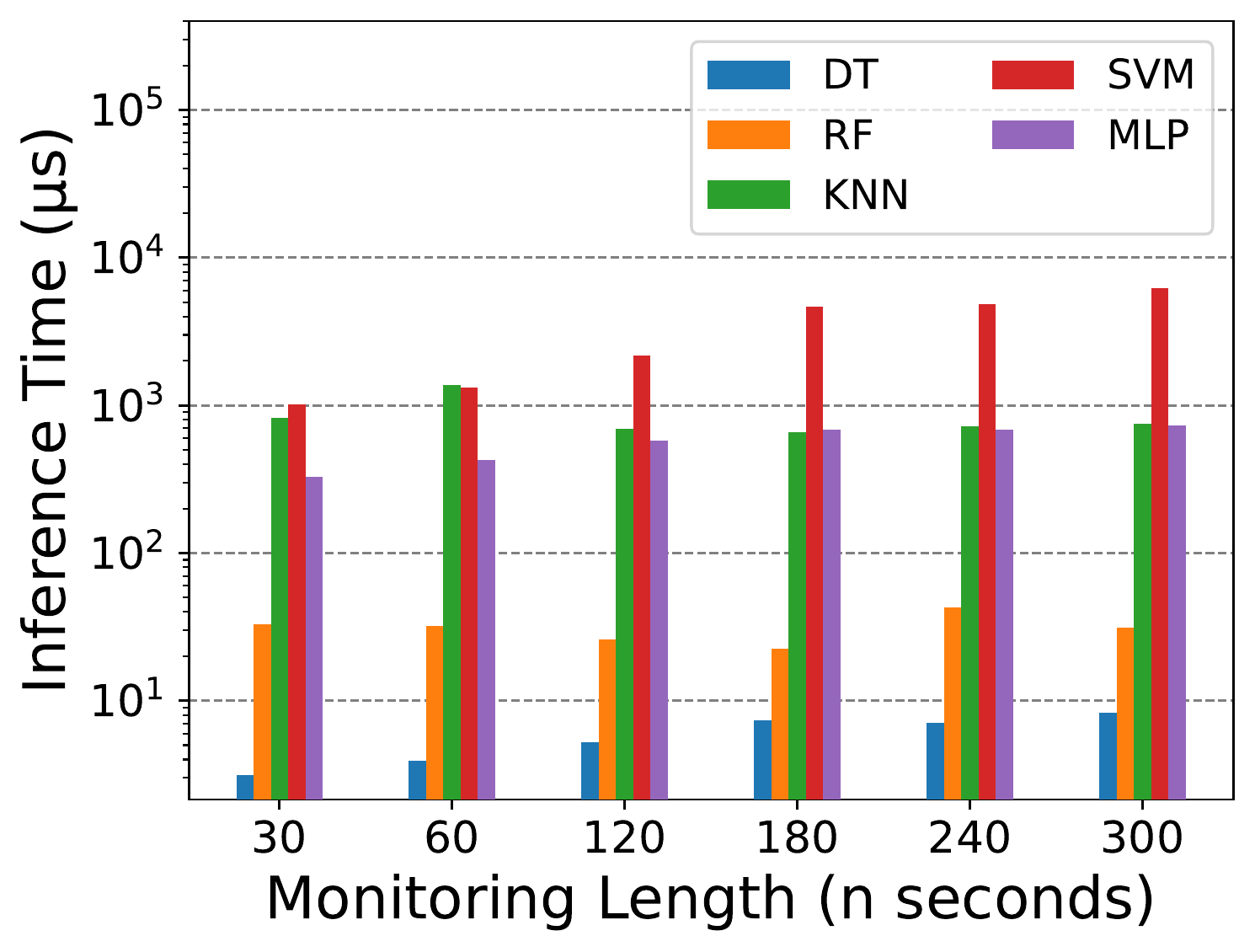}\label{fig:time_rmssd_sleep}}
  \hfill
  \subfloat[Sit.]
  {\includegraphics[width=0.33\textwidth]{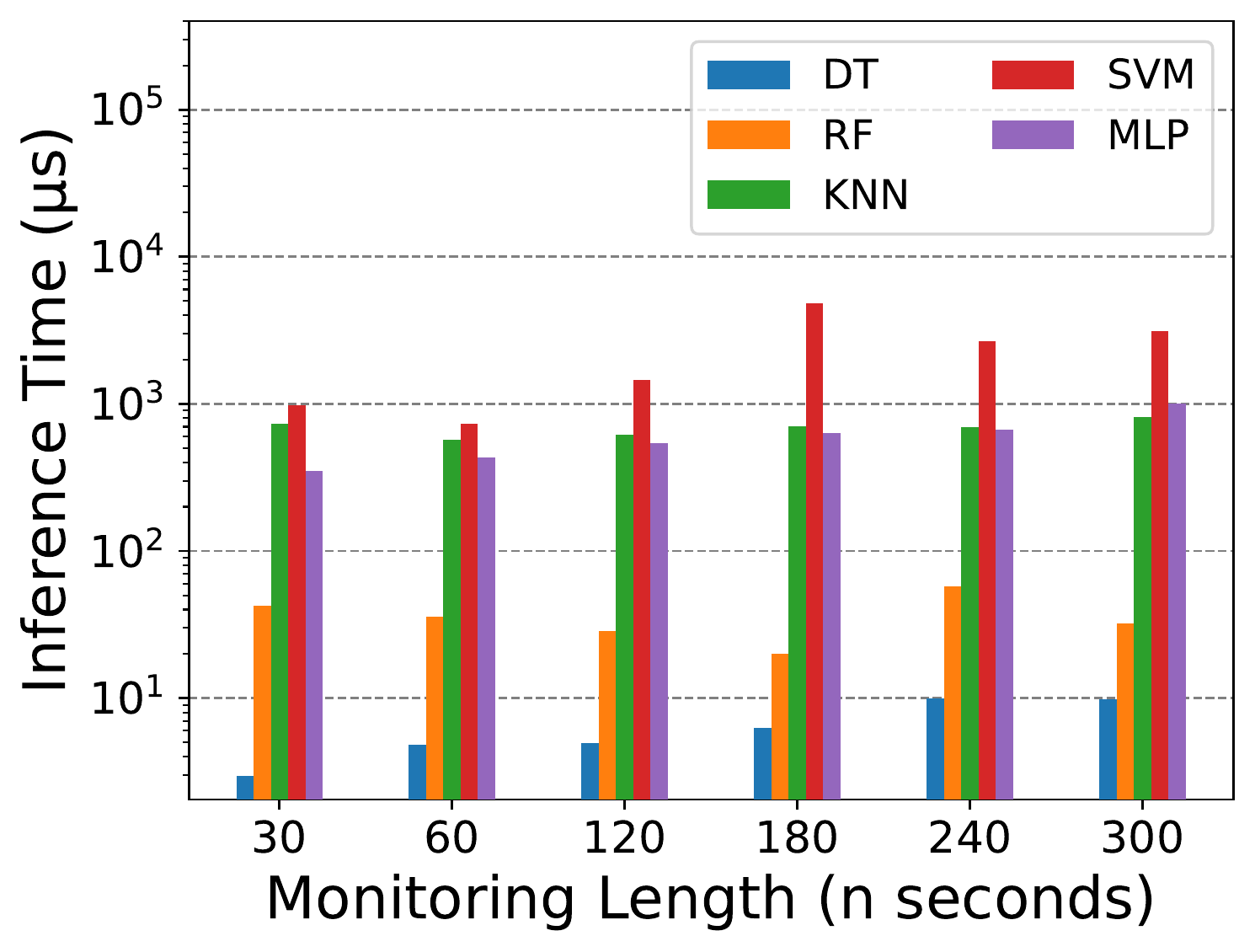}\label{fig:time_rmssd_sit}}
  \caption{ML model inference time for HRV/RMSSD estimations for different activities.}
  \label{fig:rmssd_inference_time}
\end{figure*}

\begin{figure*}
  \subfloat[Office Work.]
  {\includegraphics[width=0.33\textwidth]{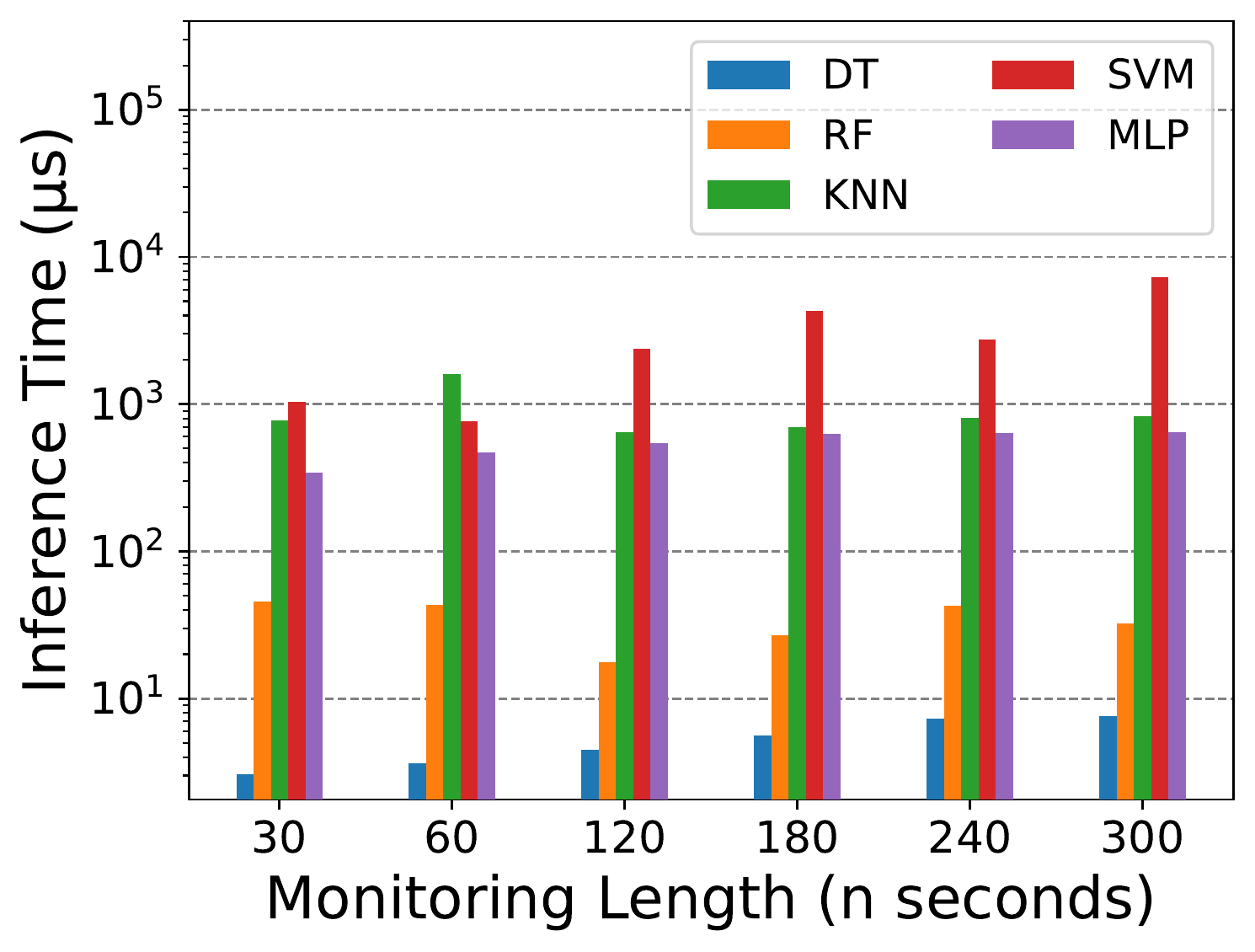}\label{fig:time_sdnn_daily}}
  \hfill
  \subfloat[Sleep.]
  {\includegraphics[width=0.33\textwidth]{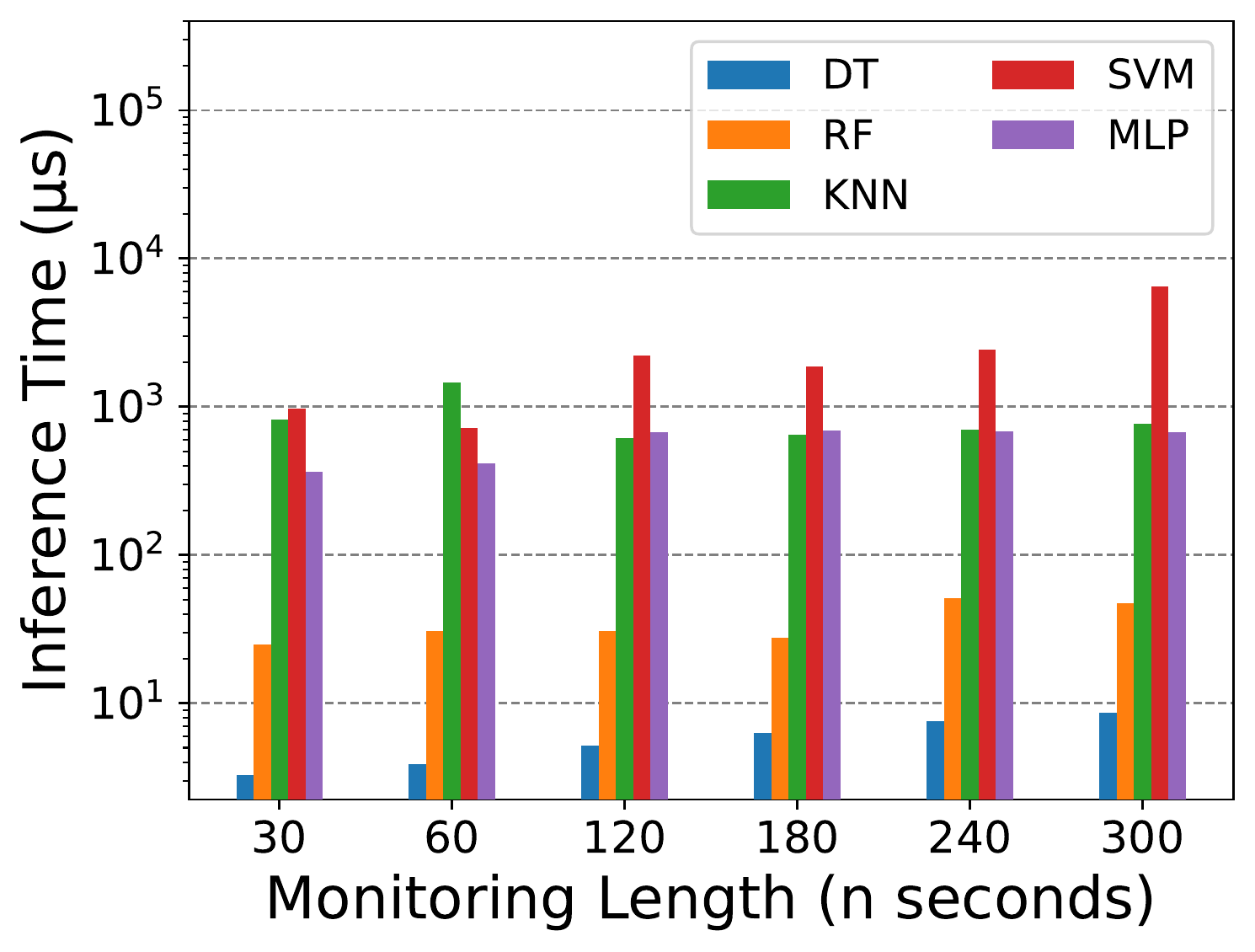}\label{fig:time_sdnn_sleep}}
  \hfill
  \subfloat[Sit.]
  {\includegraphics[width=0.33\textwidth]{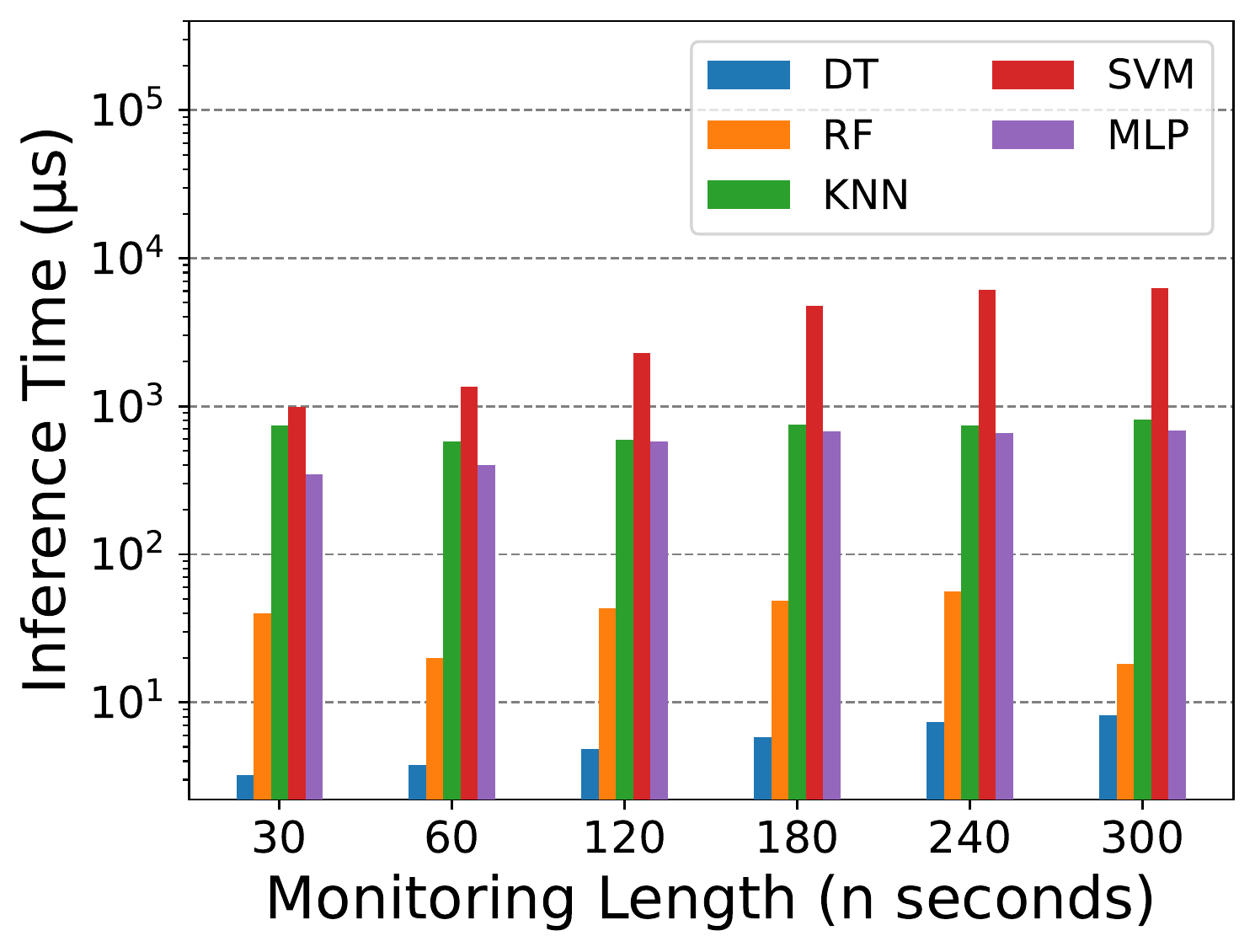}\label{fig:time_sdnn_sit}}
  \caption{ML model inference time for HRV/SDNN estimations for different activities.}
  \label{fig:sdnn_inference_time}
\end{figure*}

\subsection{Model Size Evaluation}

Figure~\ref{fig:size_rmssd} and Figure~\ref{fig:size_sdnn} give the sizes of the ML models generated by our direct and compound HRV inference method. As both figures show, the ML model sizes ranged from 2.8KB to 12.4MB. The figures also show that the model sizes are less affected by the type of activities and the HRV metric (RMSSD or SDNN). Monitoring lengths have a higher impact on the model size -- the longer the monitoring length, the more input features, and hence, larger models.

The largest factor for model size is the type of model. RF, KNN, and SVM models are usually large and approaching 10MB. 
As the input HRs to the ML models are generated from PPG signals, they are usually "messy" and lack a clear "pattern." Hence, RF/KNN/SVM requires large numbers of internal parameters to learn their "patterns." 
The MLP models, however, are smaller -- most MLP models are below 200KB, with the smallest model being 63.4KB (Figure~\ref{fig:size_rmssd_sit} at 30s) and the largest model being 468.5KB (Figure~\ref{fig:size_sdnn_daily} at 240s). DT models are even smaller, -- most DT models are below 10KB, with the smallest model being 2.8KB (Figure~\ref{fig:size_rmssd_sit} at the 30s) and the largest model is 35.4KB (Figure~\ref{fig:size_rmssd_sleep} at 120s). Considering DT and MLP models can fit in on-chip memory of hundreds of KB, they are better candidates for deploying to tiny embedded devices.

\subsection{Inference Time Evaluation}

Figure~\ref{fig:rmssd_inference_time} and Figure~\ref{fig:sdnn_inference_time} give the inference time of our compound and direct method. As both figures show, the inference time for all models was less than 7.3ms (max 7295$\mu$s). DT models are the fastest models due to their small sizes -- the inference time for DT models is between only 2.9$\mu$s to 10$\mu$s. The inference time of RF models is also fast, ranging between 17.6$\mu$s to 57.2$\mu$s. The SVM models are the slowest due to their large sizes, and their inference time ranges from 0.7ms to 7.3ms. Nonetheless, even 7.3ms is fast enough for HRV inference, indicating that all models under our methodology are fast enough for HRV monitoring with embedded devices.

The signal processing takes 65.74ms on average and 68.31ms at maximum, which is significantly slower than ML model inference.\footnote{This 65.74ms is also the average processing time for the signal only method.} Nonetheless, the signal processing is still fast enough for HRV monitoring. Note that, the total computation time for our method includes both the signal processing time and ML inference time.

\section{Related Work}
\label{sec:related_work}

Although there are many existing works on HR monitoring \cite{yuntong2022hr,bashar2019machine,chang2021deepheart,biswas2019cornet,zhang2014troika,panwar2020pp}, only a few studied HRV monitoring.



\subsection{PPG HRV Monitoring with Signal Processing}
Ghamari et al. proposed a signal processing algorithm including High-Pass and low-Pass Filters to detect R peaks~\cite{ghamari2016design}.
Srinivas et al. proposed a signal processing method including moving average and FFT to measure HRV based on the PPG wave~\cite{srinivas2007estimation}. 
However, both of them did not evaluate the accuracy of HRV estimations.
Wang et al. proposed an algorithm to estimate RR intervals from a smartwatch PPG sensor and accelerometer~\cite{wang2019s}. 
They calculated HRV measurements, such as SDNN and RMSSD. 
However, no quantitative accuracy was reported.
Blake et al. devised an HRV estimation hardware that contains a PPG sensor, an accelerometer, a Bluetooth module, and a battery \cite{blake2015development}. 
This study only compared the HRV readings from a chest strap instead of ECG with no quantitative report on accuracy.
Jankovi{\'c} and Stojanovi{\'c} designed a signal processing algorithm including a low-pass filter, Sum Slope Function, and a peak extraction function to find R peaks~\cite{jankovic2017flexible}.
They collected data for 8 minutes in the experiment and obtained one HRV for each trace, and compared them with ECG HRV. 
However, it was unclear which HRV metric they used.
Bhowmik et al. proposed an algorithm including wavelet denoising, trend removal, and peak extraction to detect R peaks in PPG signals~\cite{bhowmik2017novel}. 
They found that a 100Hz PPG sampling rate is not suitable for a smartwatch  
due to high power consumption and chose 25Hz. 
Inspired by this work, we also choose 25Hz as the PPG sampling rate to save energy. 
Saadeh et al. 
employed earlobe-attached PPG and wavelet decomposition and moving average filters to estimate RR intervals~\cite{saadeh20180}.

Note that, due to the nature of PPG light signals, the above prior studies all estimated
RR intervals first, and then converted into HRV. 
As discussed previously, inferring HRV through RR intervals can lead to low accuracy due to error amplification. Moreover, as we show in this paper, pure signal processing may not be able to handle all types of PPG noises.

\subsection{PPG HRV Monitoring with Machine Learning}

While there are prior studies that employed ML models to estimate HRV, most of these studies focused on R peak/RR interval estimation, instead of predicting HRV directly.
Everson et al. proposed a CNN-based encoder-decoder network to construct ECG waves from PPG waves and evaluated HRV based on the predicted wave \cite{everson2019biotranslator}.
They evaluated the model with the small ISPC dataset, which led to only one HRV estimation per recording. 
Similarly, Chiu et al. designed a CNN-based encoder-decoder with a sequence transformer network to generate ECG waves from PPG waves \cite{chiu2020reconstructing}.
They evaluated the model with the UQVSD dataset and the BIDMC dataset - both are datasets from barely moving patients with low motion artifacts. 
Xu et al. classified PPG signals to systolic or diastolic phase using an RNN model with the assistance of an accelerometer \cite{xu2019deep}. 
They obtained RR intervals based on the classification results and evaluated the RR interval estimations.
Wittenberg et al. compared a few CNN and GRU classification models for PPG R peaks detection \cite{wittenberg2020evaluation}. 
They classified short PPG waves based on whether the first sample in it is an R peak. 
Maritsch et al. proposed a CNN model to predict the error of the RMSSD estimations from a smartwatch~\cite{maritsch2019improving}. This work did not involve PPG signals. 
Alqaraawi et al. explored Bayesian learning to detect the PPG peaks with their collected data \cite{alqaraawi2016heart}.
However, their monitoring length was only 5 or 8 minutes, and hence, only one HRV was provided, while our data lasted 2 hours. 
Choudhury et al. used a phone camera to capture signals of a human fingertip and extract RR intervals from the collected data with adaptive neural network (ANN) and SVM \cite{choudhury2013heartsense}.
%
Instead of predicting HRV, most of the above studies mainly predicted/detected R peaks or RR intervals.
However, as we show in Section \ref{sec:motivation}, small errors in RR peak intervals can be magnified into large errors in HRV estimates.
Therefore, in this work, we estimated HRV directly (represented by SDNN and RMSSD) from PPG signals.

Commercial wearable devices may also provide HRV estimations, such as Garmin smartwatches~\cite{GraminHRV2}. However, prior work has shown that smartwatch estimations may have high errors~\cite{maritsch2019improving}. Due to their proprietary nature, we were not able to rigorously evaluate commercial wearable devices. Therefore, we focused on comparing and analyzing existing research studies in our motivation and evaluation sections.

\section{Conclusion}
\label{sec:Conclusion}



Photoplethysmography (PPG) sensors have been shown to be a good alternative for electrocardiographic (ECG) in Heart Rate Variability (HRV) monitoring. However, to be applied to practical and medical use, PPG HRV inference methods must be carefully designed. Prior work typically employed signal-processing-only or machine-learning-only methods to indirectly infer HRV from PPG signals, leading to low accuracy and large models. In this paper, we presented a compound and direct HRV inference method, which combines signal processing and machine learning to directly infer HRV. Evaluation results show that our method has errors as low as 3.5\% with model sizes of a few hundred KBs, suggesting that our method can be applied in small embedded devices and potentially for medical uses. 


\section*{Acknowledgment}
This research was in part supported by the National Science Foundation (NSF), under grants, 2155096, 2221843, 2215359, 2309760, and 2317117. Any
opinions, findings, conclusions, or recommendations expressed in this publication are those of the authors and do not necessarily reflect the view of the NSF. The authors would also
like to thank the anonymous reviewers for their insightful
comments

\bibliographystyle{ACM-Reference-Format}
\bibliography{bib-references}

\end{document}